\documentclass[conference]{IEEEtran}
\IEEEoverridecommandlockouts
\usepackage{cite}

\usepackage{xcolor}
\usepackage{algorithm}
\usepackage{algpseudocode}
\usepackage{amsmath,amssymb,amsfonts}
\usepackage{textcomp}
\usepackage{longtable}
\usepackage{subcaption}
\usepackage{ltxtable}
\usepackage{tabularx}
\usepackage{mathrsfs}
\usepackage{multicol}
\usepackage{lipsum}
\usepackage{booktabs}
\usepackage{multirow}
\usepackage{dsfont}
\usepackage{mathbbol}
\usepackage{bbm}
\usepackage{url}
\usepackage{graphicx}
\usepackage{bbding}
\usepackage{hyperref} 
\hypersetup{
    colorlinks=true,
    linkcolor=red,     
    citecolor=blue,     
    filecolor=blue,     
    urlcolor=blue       
}

\usepackage[table]{xcolor}

\definecolor{lightblue}{rgb}{0.21, 0.49, 0.74}

\usepackage{newtxtext,newtxmath} 
\newcommand{\boldres}[1]{{\textbf{#1}}}
\newcommand{\secondres}[1]{{\underline{{#1}}}}

\def\BibTeX{{\rm B\kern-.05em{\sc i\kern-.025em b}\kern-.08em
    T\kern-.1667em\lower.7ex\hbox{E}\kern-.125emX}}
\begin{document}

\title{FusAD: Time-Frequency Fusion with Adaptive Denoising for General Time Series Analysis}

\author{\IEEEauthorblockN{Da Zhang\textsuperscript{1.2}, Bingyu Li\textsuperscript{3,1}, Zhiyuan Zhao\textsuperscript{2,1}, Feiping Nie\textsuperscript{1,2}, Junyu Gao\textsuperscript{1,2\Envelope}, Xuelong Li\textsuperscript{1,2}\\}

\IEEEauthorblockA{
\textit{\textsuperscript{1}School of Artificial Intelligence, OPtics and ElectroNics (iOPEN), Northwestern Polytechnical University} \\
\textit{\textsuperscript{2}Institute of Artificial Intelligence (TeleAI), China Telecom} \\
\textit{\textsuperscript{3}Department of Electronic Engineering and Information Science, University of Science and Technology of China} \\
\{dazhang\}@mail.nwpu.edu.cn, \{libingyu0205\}@mail.ustc.edu.cn, \{tuzixini\}@163.com,\\
\{feipingnie, gjy3035\}@gmail.com, \{xuelong\_li\}@ieee.org}

\thanks{\textsuperscript{\Envelope}Corresponding author.}

}



\maketitle

\begin{abstract}

Time series analysis plays a vital role in fields such as finance, healthcare, industry, and meteorology, underpinning key tasks including classification, forecasting, and anomaly detection. Although deep learning models have achieved remarkable progress in these areas in recent years, constructing an efficient, multi-task compatible, and generalizable unified framework for time series analysis remains a significant challenge. Existing approaches are often tailored to single tasks or specific data types, making it difficult to simultaneously handle multi-task modeling and effectively integrate information across diverse time series types. Moreover, real-world data are often affected by noise, complex frequency components, and multi-scale dynamic patterns, which further complicate robust feature extraction and analysis. To ameliorate these challenges, we propose FusAD, a unified analysis framework designed for diverse time series tasks. FusAD features an adaptive time-frequency fusion mechanism, integrating both Fourier and Wavelet transforms to efficiently capture global-local and multi-scale dynamic features. With an adaptive denoising mechanism, FusAD automatically senses and filters various types of noise, highlighting crucial sequence variations and enabling robust feature extraction in complex environments. In addition, the framework integrates a general information fusion and decoding structure, combined with masked pre-training, to promote efficient learning and transfer of multi-granularity representations. Extensive experiments demonstrate that FusAD consistently outperforms state-of-the-art models on mainstream time series benchmarks for classification, forecasting, and anomaly detection tasks, while maintaining high efficiency and scalability. Code is available at \href{https://github.com/zhangda1018/FusAD}{FusAD}.

\end{abstract}

\begin{IEEEkeywords}
Time Series Analysis, Time-Frequency Fusion, Adaptive Denoising
\end{IEEEkeywords}

\section{Introduction}

\textcolor{black}{Time series analysis (TSA) is fundamental to forecasting dynamic systems \cite{yue2022ts2vec}, with critical applications in weather \cite{wu2021autoformer}, healthcare \cite{chen2025impact}, and energy \cite{turowski2024generating}.}
By modeling historical dynamics and predicting future trends using time series, it not only improves the timeliness of fault warnings and risk mitigation, but also directly drives the development of intelligent applications such as personalized recommendation \cite{fan2024recommender}, automated decision-making \cite{Position}, and medical anomaly detection \cite{xuanomaly}. 
With advances in sensor technologies and data acquisition, modern time series exhibit new characteristics including multi-modality, high noise levels, and task diversity \cite{liu2024time}, posing unprecedented challenges to the generalization and adaptability of time series modeling.

Recently, deep learning-based approaches have significantly advanced the development of TSA \cite{cheng2025convtimenet}. 
Early neural architectures such as RNNs \cite{lu2024trnn, piao2025garnn}, TCNs \cite{luo2024moderntcn, dissanayaka2024temporal}, and LSTMs \cite{chen2024pioneering} have achieved remarkable progress in dynamic modeling and pattern recognition. 
Notably, Transformer-based models (e.g., Informer \cite{zhou2021informer}, PatchTST \cite{nietime}, iTransformer \cite{liuitransformer}), leveraging self-attention mechanisms, excel at capturing long-term temporal dependencies, leading to both performance improvements and scalability innovations across major tasks ranging from energy forecasting and weather prediction to equipment health monitoring \cite{wen2023transformers, woo2024unified, liu2024generative}. 
Additionally, the deep and multi-layered representation capabilities of these networks provide inherent advantages for learning from large-scale and complex time series data, laying a solid foundation for the development of universal models and broader multi-scenario extensions \cite{liang2024foundation}.

\begin{figure}[t]
    
\includegraphics[width=0.49\textwidth]{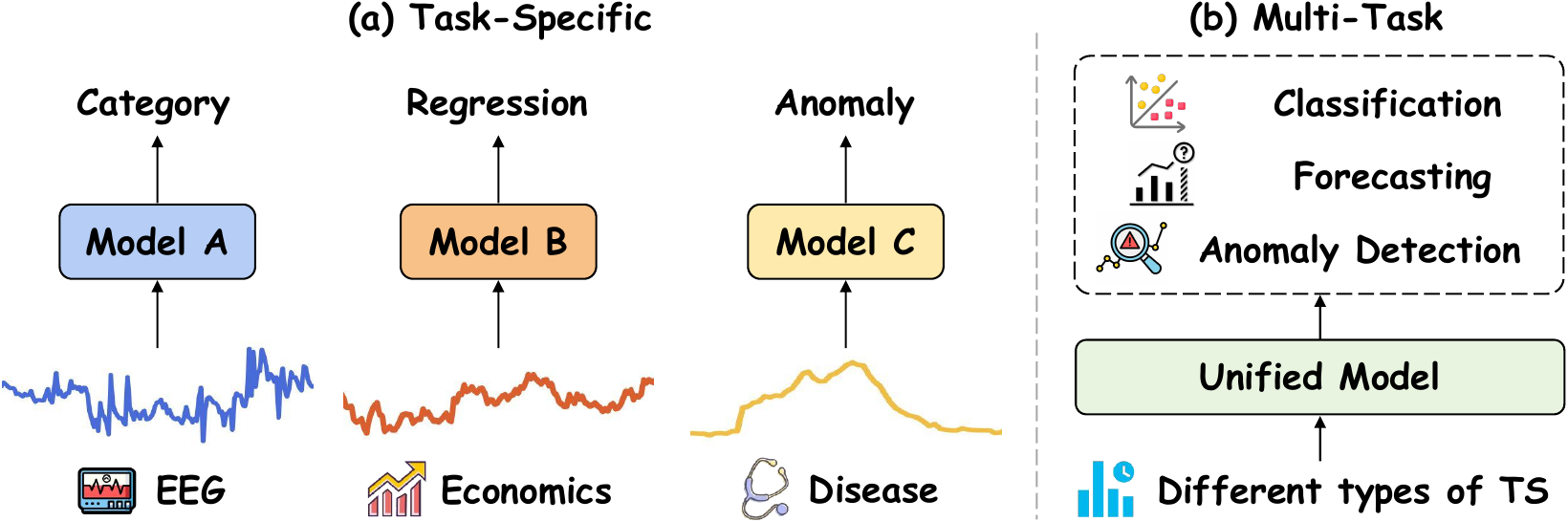}
    \caption{
    (a) Task-specific models are trained separately in time series domains with significant distribution differences. \textcolor{black}{For example, fluctuating Electroencephalogram (EEG) signals are used for recognition domains,} while economic data such as exchange rates often require prediction. Disease data such as seasonal influenza typically exhibit sudden anomalies over longer time periods. 
    (b) The proposed unified framework processes time series data from different domains, utilizing a common representation to achieve unified multi-task generalization.}
    \vspace{-0.5cm}
    \label{fig1}
\end{figure}

Despite these advancements, most existing methods follow the paradigm of ``one model per task/scene", training dedicated networks for specific problems or data domains \cite{gao2024units}. 
As a result, each model must be separately optimized for a given target, lacking the capacity to deeply exploit common knowledge across tasks or directly transfer to different scenarios \cite{liu2024unitime}. 
As a result, the concepts of universal representation learning and unified multi-task modeling have emerged (e.g., TS2Vec \cite{yue2022ts2vec}, TimesNet \cite{wutimesnet}, TVNET \cite{litvnet}), utilizing techniques such as self-supervised multi-scale encoding \cite{zhang2024self}, contrastive learning \cite{liu2024timesurl}, or masked pre-training \cite{cheng2023timemae} to improve transferability and robustness across diverse tasks within a common framework. 
Figure \ref{fig1} shows examples of these two paradigms. 
Nevertheless, current approaches remain far from resolving the core challenges of a unified analysis framework, facing three prominent issues:

\begin{itemize}
\item \textcolor{black}{\textbf{Insufficient multi-scale and spectral characterization}. 
Time series often encompass long-term trends (e.g., seasonality), short-term fluctuations (e.g., anomalies), and non-stationary dynamics (e.g., changing statistical properties), mixing various frequency components \cite{liu2023koopa, dai2024ddn, fan2024deep}.}
\item \textbf{Lack of robustness in noisy and anomalous environments}. Time series data are typically subject to noise, outliers, and missing values. Traditional static denoising methods struggle to adapt to unknown noise types and complex anomalies, leading to confused feature extraction and degraded performance \cite{chengrobusttsf}.
\item \textbf{Limited model efficiency and scalability}. When faced with large-scale, high-dimensional, or long sequence data, many models incur high computational and memory costs, resulting in slow training and inference \cite{ni2025timedistill}. In addition, complex structures and large parameter sizes hinder adaptation to new tasks or scaling to larger datasets \cite{hansofts}.
\end{itemize}

\begin{figure}[h]
    
    \includegraphics[width=0.49\textwidth]{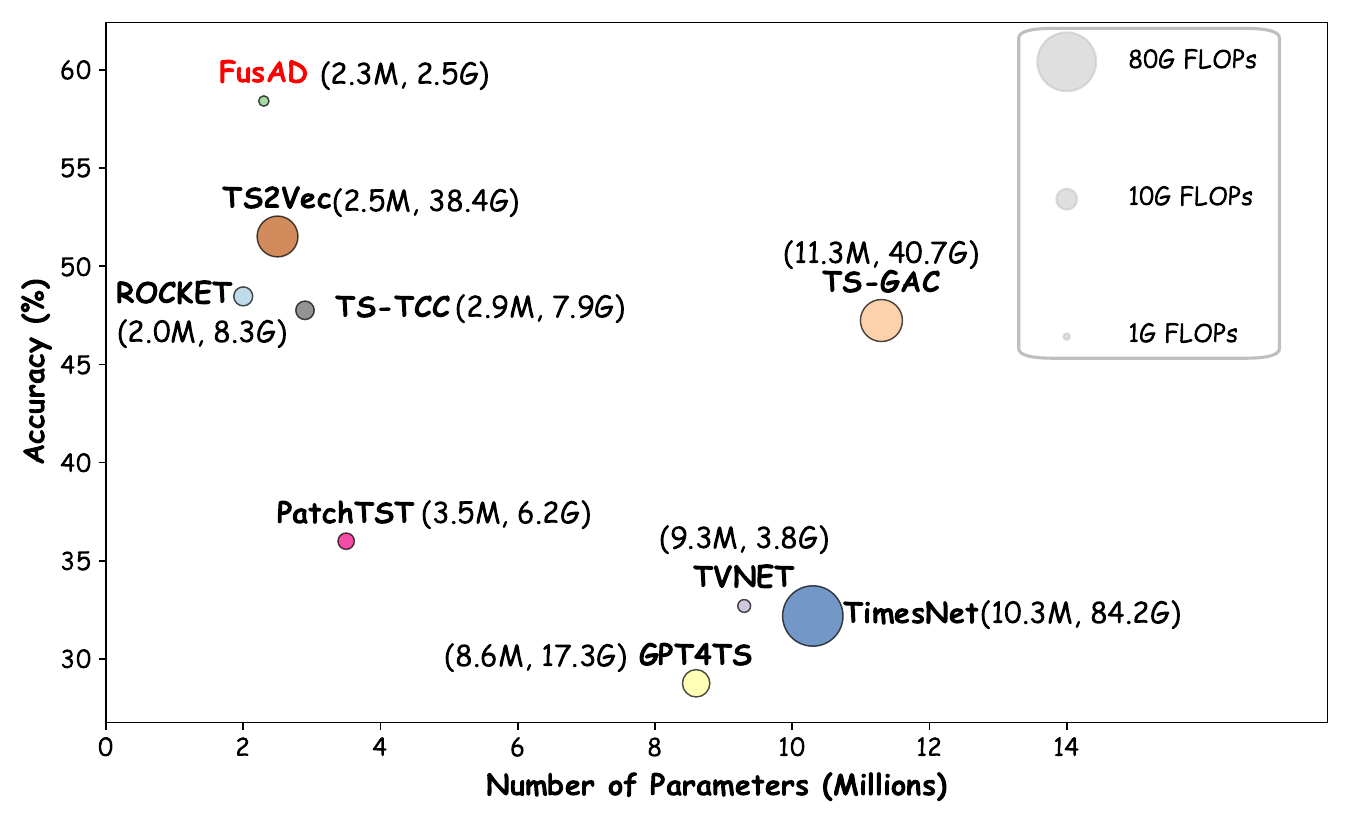}
    \caption{
    \textcolor{black}{A comparison of FusAD's performance on the Handwriting classification dataset (from UCR archive) \cite{dau2019ucr}. The plot shows model accuracy relative to the number of parameters (in Millions, M) and the computational cost (in GigaFLOPS, G). FusAD demonstrates superior accuracy with significantly lower parameter and computational overhead compared to baseline models.}
    }
    \label{fig2}
\end{figure}

To address these challenges, we present FusAD, a unified TSA framework that integrates time-frequency feature modeling with adaptive denoising, aiming for efficient and multi-task TSA. 
Specifically, FusAD introduces an adaptive time-frequency fusion module, co-embedding both Fourier and Wavelet transforms to efficiently capture multi-scale dynamics such as global periodicity and local abrupt changes. 
It further employs an adaptive thresholding mechanism for feature enhancement and noise suppression across frequency bands, thus boosting robustness at the sequence level. 
In addition, FusAD features information fusion and generic decoding components, enabling deep integration of multi-source heterogeneous time series via inter-layer interaction, and generating multi-granular semantic outputs for various tasks. 
Coupled with masked pre-training techniques \cite{zhang2023trid}, FusAD promotes the learning and transfer of unified representations in multi-task scenarios. 
Extensive experiments demonstrate that FusAD achieves state-of-the-art (SOTA) performance on multiple tasks, including classification, forecasting, and anomaly detection, and surpasses existing methods in both parameter efficiency and scalability (as shown in Figure \ref{fig2}).
The main contributions of this paper are as follows:

\begin{itemize}
    \item We propose FusAD, a novel unified TSA framework that integrates time-frequency modeling and adaptive denoising, effectively addressing multi-scale dynamic modeling and robustness under complex noise conditions.
    \item We design an innovative time-frequency fusion module that combines Fourier and Wavelet transforms, enabling efficient extraction of both global periodic and local abrupt features, while applying adaptive thresholding to enhance representations and noise resilience across diverse frequency bands.
    \item We develop an information fusion and universal decoding component, which enables deep integration of multi-source heterogeneous time series and unified multi-task modeling via inter-layer interaction and masked pre-training, significantly improving knowledge sharing and representation generalization across tasks.
    \item Experimental results show that FusAD achieves superior performance in forecasting, classification, and anomaly detection tasks, and outperforms existing baselines in parameter efficiency and scalability.
\end{itemize}
\vspace{-0.4cm}

\textcolor{black}{The remainder of this paper is organized as follows. Section \ref{related work} provides a review of related work. Section \ref{method} details FusAD's framework, including its architecture, different components and training strategy. Section \ref{experiment} presents the results of extensive experiments across multiple tasks. Section \ref{model analysis} offers a deeper analysis of various evaluations. Finally, Section \ref{conclusion} concludes the paper and discusses future research directions.}

\section{Related Work}
\label{related work}

\subsection{Time Series Analysis} 
\textcolor{black}{Classical methods like ARIMA \cite{shumway2017arima} and Holt-Winters models \cite{tratar2016comparison} are interpretable but fail on complex, non-linear, or high-dimensional data \cite{zhang2024multivariate}.
Recently, deep learning methods, especially \textcolor{black}{Multi-Layer Perceptron (MLP)-} \cite{yeh2024rpmixer, ekambaram2023tsmixer} and Transformer-based models \cite{zhou2021informer, nietime, liuitransformer}, have made significant progress.}
MLP-based models are gaining traction due to their low computational complexity \cite{challu2023nhits}. 
For instance, DLinear \cite{zeng2023transformers} achieves efficient prediction by combining decomposition techniques with MLPs. 
However, current MLP models struggle to capture periodicity and inter-variate relationships \cite{litvnet}. 
Meanwhile, Transformer-based models exhibit strong performance by leveraging self-attention mechanisms to capture long-term dependencies and critical global information \cite{wu2021autoformer, chen2024pathformer}. 
Whereas, their scalability and efficiency are constrained by the quadratic complexity of the attention mechanism. 
\textcolor{black}{Subsequent works like Pyraformer \cite{liupyraformer} and 
PatchTST \cite{nietime} try to mitigate this by designing more efficient attention or patching strategies. Despite these innovations, most works remain task-specific and lack a unified mechanism to extract global and local features. Moreover, they primarily model 1D temporal variations and still face computational challenges in long-term forecasting.}

\subsection{Unified Multi-task Modeling} 

Recent research targets unified multi-task frameworks for large-scale time series \cite{fraikin2024t, liu2024time}.
Methods such as TS2Vec \cite{yue2022ts2vec}, TimesNet \cite{wutimesnet}, GPT4TS \cite{zhou2023one}, and TVNET \cite{litvnet} learn general-purpose embeddings through self-supervised or contrastive learning objectives, aiming for universality in classification \cite{zhang2024multivariate}, forecasting \cite{liuitransformer}, and anomaly detection \cite{zamanzadeh2024deep} tasks. 
Some approaches, such as Uni2TS \cite{woo2024unified}, explore foundation models or prompt-based paradigms to apply large-scale pretraining and transfer learning across broad time series domains. 
Despite these advances, most research still primarily focuses on temporal structures, seldom exploiting frequency information or domain-specific noise \cite{zhang2025beyond}.
Furthermore, most universal frameworks rely on pretraining tasks (such as masked modeling \cite{cheng2023timemae} or augmented contrastive learning \cite{cheng2025convtimenet}) to achieve generalization, yet still adopt static feature extraction backbones and lack the capacity to adaptively fuse or denoise information under varying noise and spectral conditions \cite{dong2024timesiam}. 
In contrast, FusAD combines generic semantic modeling with dynamic adaptive time-frequency fusion, delivering a more comprehensive and sequence-aware approach for multi-task TSA and robust transfer across data sources through the collaborative learning of Fourier and wavelet features, together with frequency-wise adaptive denoising.

\subsection{Time-Frequency Feature Learning}

Spectral analysis is increasingly recognized as crucial for robust time series modeling \cite{zhang2025beyond}.
Models utilizing Fourier transforms (such as FiLM \cite{zhou2022film}, Non-stationary Transformer \cite{liu2022non}, FAN \cite{yefrequency}) and Wavelet decompositions (such as WaveletMixer \cite{zhang2025waveletmixer} and WPMixer \cite{murad2025wpmixer}) have shown success in capturing multi-scale dynamics (trends, cycles, abrupt changes). 
\textcolor{black}{Further emphasizing the power of frequency-centric design, FITS \cite{xufits} introduces a remarkably lightweight model that reinterprets forecasting as an interpolation task within the complex frequency domain. In parallel, Graph WaveNet \cite{wu2019graph} have proven effective at capturing long-range temporal dependencies in spatial-temporal data by stacking dilated convolutions.}
However, existing work often employs fixed spectral filters or naive denoising, lacking mechanisms to adaptively enhance relevant components and suppress non-stationary noise, especially in scenarios with distribution shifts, missing values, or changes in signal-to-noise ratio (SNR) \cite{zhou2024denoising}. 
\textcolor{black}{In contrast, our work introduces a learnable time-frequency fusion module, which combines Fourier and wavelet transforms with an adaptive thresholding mechanism to dynamically denoise and enhance features, enabling robust learning in complex environments.}

\section{Method}
\label{method}

\subsection{Problem Definition}
A multivariate time series consists of $N$ univariate time series, each of length $T$, represented as $\mathbf{X} \in \mathbb{R}^{N \times T}$. The dataset contains $k$ instances $\mathcal{D} = \{(\mathbf{X}_i, \mathbf{y}_i)\}_{i=1}^k$.
We aim to address the following three tasks in a unified framework:
\begin{itemize}
\item \textbf{Classification:} Predict the class label $y^{(c)}_i \in \{1, 2, \ldots, l\}$ for each instance $\mathbf{X}_i$.

\item \textbf{Forecasting:} Predict the future $h$ time steps based on the observed history, i.e., output $\hat{\mathbf{X}}_{i}^{(f)} \in \mathbb{R}^{N \times h}$ given $\mathbf{X}_i$.

\item \textbf{Anomaly detection:} Identify the anomaly score/vector $a^{(a)}_{i,t}$ for each time point $t$ in $\mathbf{X}_i$.
\end{itemize}
Formally, our framework jointly learns mapping functions for all three tasks:
\begin{equation}
(\hat{y}^{(c)}_i, \hat{\mathbf{X}}_{i}^{(f)}, \hat{\mathbf{a}}^{(a)}_{i,1:T}) = \mathcal{F}_{multi}(\mathbf{X}_i; \Theta)
\end{equation}
where $\mathcal{F}_{multi}$ is the multi-task model. The goal is to optimize the sum of task-specific objectives, demonstrating the robust and generalizable nature of our approach.

\begin{figure*}[t]
    \centering
    \includegraphics[width=0.99\textwidth]{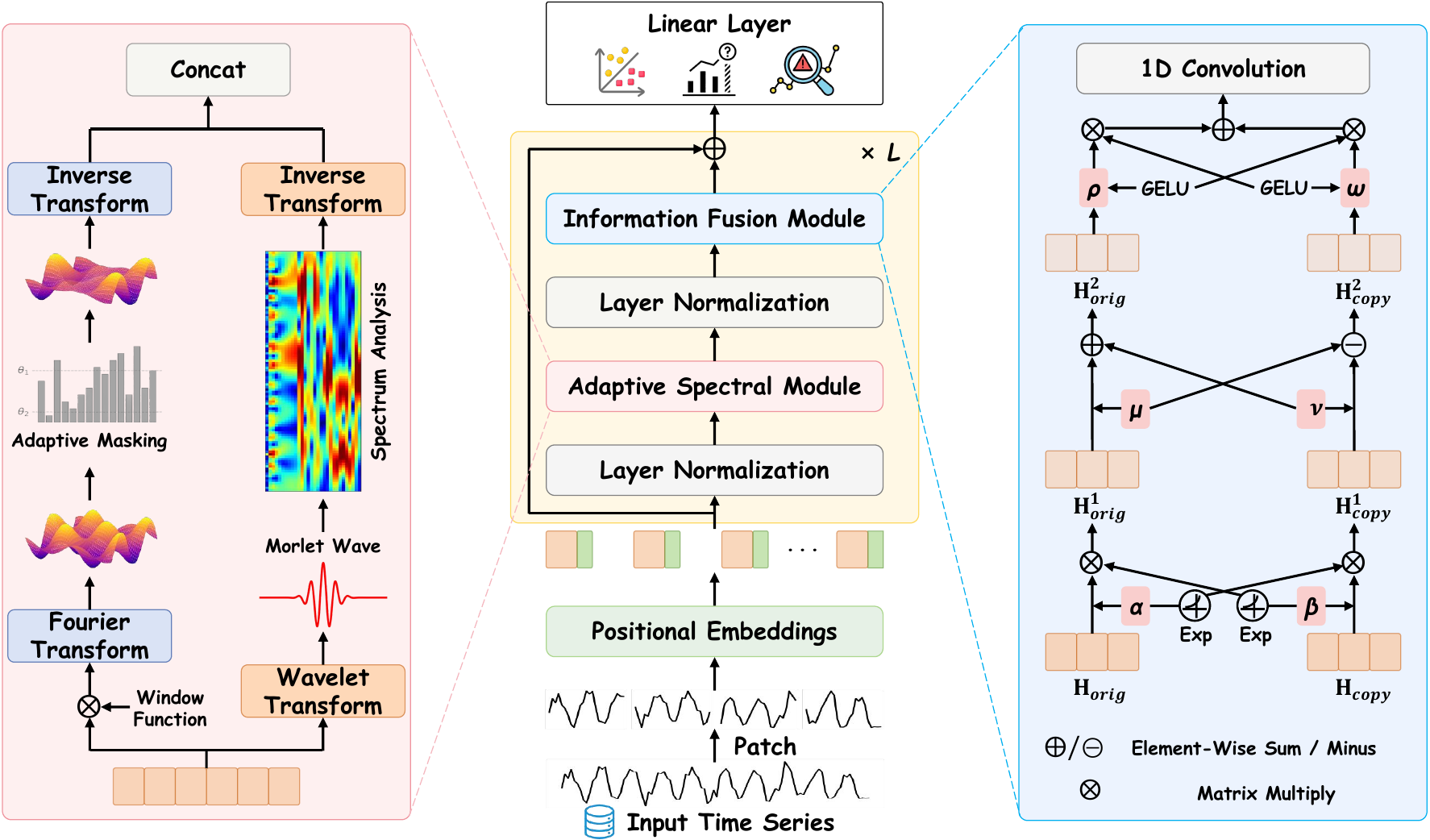}
    \caption{
        Overall architecture of FUsAD. 
        Different types of input time series are divided into multiple patches with positional embeddings added (represented as one-dimensional sequences for convenience). 
        These input embeddings first pass through an ASM, which utilizes Fourier and Wavelet transforms for time-frequency conversion analysis and applies adaptive thresholding for noise reduction. 
        Subsequently, the transformed features are sent to the IFM, which refines the features through interactive convolution and activation functions. 
        Finally, the features processed through multiple layers are fed into the linear layer for the final multiple task.
    }
    \vspace{-0.5cm}
    \label{fig3}
\end{figure*}

\subsection{Overview Framework}

\textcolor{black}{The overall architecture of our proposed FusAD framework is illustrated in Figure 3. The model is designed as a scalable hierarchical structure consisting of $L$ identical layers to progressively refine the time series representation. Each layer contains an Adaptive Spectral Module (ASM) and an Information Fusion Module (IFM), with Layer Normalization (LN) applied before each module to stabilize training. This layered design allows the model to learn features at different levels of abstraction.}

\textcolor{black}{The data processing pipeline begins with an embedding layer. The input time series is first partitioned into non-overlapping patches to capture local temporal information. Each patch is projected into feature space, and positional embeddings are added to retain sequential order. This embedded representation is used as the initial input to the first layer of the FusAD hierarchy. The representation then passes through the stack of $L$ layers. Within each layer, ASM is the first to process the features. The ASM performs comprehensive time-frequency analysis, utilizing parallel Fourier and Wavelet transforms to decompose the sequence, capturing global periodic patterns and local transient characteristics. This module includes an adaptive denoising mechanism that learns to filter out irrelevant frequency components, enhancing robustness. Following the ASM, features are fed into IFM. The IFM models inter-dependencies across different variates of the time series. It uses an interactive structure with multiple convolutional pathways and activation functions to merge and refine multivariate information, ensuring cross-variate relationships are captured. After processing through all $L$ layers, the final representation is passed to a linear layer that projects the features for specific downstream tasks. Depending on the task, a corresponding head produces the final output, such as classification, forecasting, or anomaly detection. This unified structure enables FusAD to learn generalizable representations suitable for diverse TSA tasks while maintaining high efficiency. Algorithm \ref{algorithm1} provides the detailed pseudocode for this process.}

\begin{algorithm}[t]
\caption{\textcolor{black}{FusAD: Time-Frequency Fusion with Adaptive Denoising}}
\textbf{Input:} Multivariate time series $\mathbf{X} \in \mathbb{R}^{N\times T}$ \\
\textbf{Output:} Classification result $\hat{\mathbf{y}}^{(c)}$, Prediction $\hat{\mathbf{X}}^{(f)}$, Anomaly score $\hat{\mathbf{a}}^{(a)}$
\begin{algorithmic}[1]
\State Divide $\mathbf{X}$ into patches $\{\mathbf{B}_1, \dots, \mathbf{B}_Z\}$ \Comment{Patch embedding}
\For{each patch $\mathbf{B}_i$}
    \State $\mathbf{B}_i' \gets \mathrm{Conv1D}(\mathbf{B}_i)$ 
    \State $\mathbf{B}_i^E \gets \mathbf{B}_i' + \mathbf{E}_i$ \Comment{Add positional encoding}
\EndFor
\For{$l = 1$ to $L$}
    \State \textbf{// Adaptive Spectral Module (ASM)}
    \State $\mathbf{F} \gets \mathrm{FFT}(\mathrm{Hanning}(\mathbf{B}^E))$ \Comment{Fourier transform}
    \State $P \gets |\mathbf{F}|^2$ \Comment{Power spectrum}
    \State $\mathbf{F}' \gets \mathbf{F} \otimes [\theta_1 \leq P \leq \theta_2]$ \Comment{Adaptive denoising}
    \State $\mathbf{W} \gets \mathrm{CWT_{Morlet}}(\mathbf{B}^E)$ \Comment{Wavelet transform}
    \State $\mathbf{H} \gets \text{Concat}(\mathrm{IFFT}(\mathbf{F}'), \mathrm{ICWT}(\mathbf{W}))$
    \State \textbf{// Information Fusion Module (IFM)}
    \State Duplicate $\mathbf{H}$ as $\mathbf{H}_{\text{orig}}$, $\mathbf{H}_{\text{copy}}$
    \State $\mathbf{H}^1_{\text{orig}} \gets \mathbf{H}_{\text{orig}} \otimes \exp(\beta(\mathbf{H}_{\text{copy}}))$
    \State $\mathbf{H}^1_{\text{copy}} \gets \mathbf{H}_{\text{copy}} \otimes \exp(\beta(\mathbf{H}_{\text{orig}}))$
    \State $\mathbf{H}^2_{\text{orig}} \gets \mathbf{H}^1_{\text{orig}} + \nu(\mathbf{H}^1_{\text{copy}})$
    \State $\mathbf{H}^2_{\text{copy}} \gets \mathbf{H}^1_{\text{copy}} - \mu(\mathbf{H}^1_{\text{orig}})$
    \State $\mathbf{H}^3_{\text{orig}} \gets \rho(\mathbf{H}^2_{\text{orig}}) \otimes \mathrm{GELU}(\omega(\mathbf{H}^2_{\text{copy}}))$
    \State $\mathbf{H}^3_{\text{copy}} \gets \omega(\mathbf{H}^2_{copy}) \otimes \mathrm{GELU}(\rho(\mathbf{H}^2_{\text{orig}}))$
    \State $\mathbf{H}_{\text{IFM}} \gets \mathrm{Conv1D}(\mathbf{H}^3_{\text{orig}} + \mathbf{H}^3_{\text{copy}})$ \Comment{Layer output}
\EndFor
\State \textbf{// Multi-task decoding}
\State $\hat{\mathbf{y}}^{(c)} \gets \mathrm{ClsHead}(\mathbf{H}_{\text{IFM}})$; $\hat{\mathbf{X}}^{(f)} \gets \mathrm{ForeHead}(\mathbf{H}_{\text{IFM}})$; $\hat{\mathbf{a}}^{(a)} \gets \mathrm{AnoHead}(\mathbf{H}_{\text{IFM}})$
\State \Return $\hat{\mathbf{y}}^{(c)},~\hat{\mathbf{X}}^{(f)},~\hat{\mathbf{a}}^{(a)}$
\end{algorithmic}
\label{algorithm1}
\end{algorithm}

\subsection{Embedding Layer}

Given an input time series dataset $\mathcal{D}$, where each sample \(\mathbf{X} \in \mathbb{R}^{N \times T}\) has \(N\) dimensions and length \(T\). First, the sample $\mathbf{X}$ is divided into a set of \(Z\) patches \(\{B_1, B_2, \ldots, B_Z\}\), where each patch \(B_i\) captures a segment of \(\mathbf{X}\). The dimension of each patch is determined by a predefined patch size \(b\), so each patch \(B_i \in \mathbb{R}^{N \times b}\). Then, each patch is mapped to a new dimension \(b'\) through a one-dimensional convolutional layer, i.e.,
\begin{equation}
B_i \xrightarrow{\text{Conv1D}} B_i' \in \mathbb{R}^{N \times b'}.
\end{equation}
Next, for each patch \(B_i'\), position embeddings are added to preserve the temporal order disrupted during segmentation. The position embedding for the \(i\)-th patch is denoted as \(E_i\), with the same dimension as the patch. The final position embedding obtained is
\begin{equation}
B^E_i = B_i' + E_i.
\end{equation}
Thus, we obtain \(B^E = \{B^E_1, B^E_2, \ldots, B^E_Z\}\), and these embedding vectors are learnable.

\subsection{Adaptive Spectral Module}

Inspired by WFTNet \cite{liu2024wftnet}, we propose an Adaptive Spectral Module (ASM) for time-frequency processing. 
Features processed by the normalization layer are analyzed through different branches: one branch employs the Fourier Transform to capture global periodic patterns, while another uses the Wavelet Transform to capture local patterns. 
Additionally, during the Fourier Transform, adaptive filtering is applied to attenuate the high-frequency and low-frequency components of the sequence.

\subsubsection{\textcolor{black}{Discrete Fourier Transform (DFT)}}

DFT is a method that converts time-domain sequence into frequency-domain representations, allowing the sequence to be decomposed into a series of components with different frequencies. 
For a discrete time series \( f[h] = [f_0, f_1, \ldots, f_{H-1}] \) of length \( H \), where \( 0 \leq h \leq H-1 \), the one-dimensional DFT transforms it into frequency-domain characteristics:

\begin{equation}
F[k] = \sum_{h=0}^{H-1} f[h]e^{-j\frac{2\pi}{H}kh} = \sum_{h=0}^{H-1} f[h]W_H^{kh},
\end{equation}
\noindent where each component \( F[k] \) is a complex number that includes amplitude and phase information. \( j \) denotes the imaginary unit, and \( W_N = e^{-j\frac{2\pi}{H}} \) is the base frequency of the \textcolor{black}{Discrete Fourier Transform}. 
The purpose of DFT is to identify global periodic characteristics from the sequence, and its discrete nature is highly compatible with digital signal processing. 
To improve computational efficiency, the Fast Fourier Transform (FFT) is commonly used as a faster method to compute DFT.

For the feature sequence \( B^E \) processed by the LN, we initially apply a Hanning Window for preprocessing. The Hanning window smooths the sequence by gradually reducing the weight at the beginning and end of the sequence, effectively mitigating boundary effects. 
Subsequently, we obtain its frequency-domain representation \textbf{F} by performing FFT in the spatial dimension:

\begin{equation}
\textbf{F} = \mathcal{F}\mathcal{F}[\text{Hanning}(B^E)] \in \mathbb{C}^{N \times T'},
\end{equation}
\noindent
where \( \mathcal{F}\mathcal{F}[\cdot] \) represents the one-dimensional FFT operation, and \( T' \) is the length of the transformed sequence in the frequency domain, which is generally different from \( T \) and depends on the characteristics of the FFT implementation and the data. 
Transforming each variate of the time series independently results in a comprehensive frequency-domain representation \textbf{F}, which contains the spectral characteristics of the original time series across all variates.

\subsubsection{Adaptive Denoising}
High-frequency noise typically manifests as rapid fluctuations, which often deviate from the underlying trend. 
Low-frequency components may contain trend distortions, especially when the dataset exhibits long-term non-linear trends. 
Thus, we design an adaptive filter that dynamically adjusts threshold values based on the characteristics of different dataset to remove high and low-frequency components, thereby enhancing the robustness of the model.
\textcolor{black}{This adaptive capability is particularly crucial for non-stationary datasets or those with less distinct periodic structures, as it allows the model to learn the appropriate frequency band to retain, thereby preventing the erroneous removal of meaningful, aperiodic signal components.}

After obtaining the frequency domain representation \textbf{F}, we calculate its power spectrum \textbf{P}, expressed as \( \textbf{P} = |\textbf{F}|^2 \), which provides a measure of the strength of different frequencies in sequence.
The core of denoising lies in adaptively filtering out high and low-frequency components from the power spectrum \textbf{P}. 
By setting two learnable threshold values in training phase, \( \theta_1 \) and \( \theta_2 \), the filter can dynamically adjust according to the spectral characteristics of the dataset. The frequency domain representation \textbf{F}$'$ after denoising is given by:
\begin{equation}
\textbf{F}' = \textbf{F} \otimes  [\theta_1 \leq \textbf{P} \leq \theta_2],
\end{equation}
\noindent where \( \otimes  \) denotes the Hadamard Product, and \( [\theta_1 \leq \textbf{P} \leq \theta_2] \) is a Boolean mask that ensures only frequencies within the range \( \theta_1 \) and \( \theta_2 \) are retained.
By adaptively selecting the threshold values \( \theta_1 \) and \( \theta_2 \), ASM can efficiently denoise while preserving critical information, enhancing the model's effectiveness.

\subsubsection{\textcolor{black}{Continuous Wavelet Transform (CWT)}}

CWT decomposes sequences into time and frequency domains, and it is more effective than DFT in capturing the local periodic structure of sequences. 
Moreover, due to the characteristics of wavelet transformation, it can also achieve noise reduction effectively.

For a time sequence \( f[h] = [f_0, f_1, \ldots, f_{H-1}] \) of length \( H \), where \( 0 \leq h \leq H-1 \), the definition of CWT is as follows:
\begin{equation}
CWT(\sigma, \tau) = \int f_h \cdot \frac{1}{\sqrt{\sigma}} \cdot \Psi^*\left(\frac{h-\tau}{\sigma}\right) dh \quad (0 \leq h < H),
\end{equation}
\noindent where \( \sigma \) and \( \tau \) represent scale factor and translation factor, and \( \Psi(t) \) denotes the selected mother wavelet function, with \( * \) indicating complex conjugation. The relationship between \( \Psi \) and \( \Psi^* \) is: \( \Psi^*_{\sigma, \tau}(h) = \frac{1}{\sqrt{\sigma}} \Psi\left(\frac{h-\tau}{\sigma}\right) \).

Due to the excellent performance of Morlet wavelets in time-frequency analysis of signals \cite{zhang2025waveletmixer}, they are used as the mother wavelet function and implemented through convolution. 
Subsequently, we obtain its frequency-domain representation $\mathbf{W}$ by performing CWT in the temporal dimension:
\begin{equation}
\textbf{W} = CWT_{\text{Morlet}}(B^E) \in \mathbb{C}^{N' \times T},
\end{equation}
\textcolor{black}{Finally, the features from both branches are transformed back to the time domain using the Inverse Fast Fourier Transform (IFFT) and Inverse Continuous Wavelet Transform (ICWT) respectively, and then concatenated to form the aggregated feature $\textbf{H}$:}
\begin{equation}
\textbf{H} = Concat(IFFT(\textbf{F}'),ICWT(\textbf{W})),
\end{equation}

\subsection{Information Fusion Module}
After ASM for denoising and feature enhancement, we propose an interactive Information Fusion Module (IFM), which utilizes a multi-layer convolution-activation structure as shown in Figure \ref{fig3}. The design of IFM includes parallel convolution with different kernel sizes and activation functions at different positions, aiming to merge information from different variates through interactive fusion.

Specifically, we duplicate the input features \( \textbf{H}_{\text{orig}} \)  to obtain \( \textbf{H}_{\text{copy}} \), and use different convolutional kernels to extract features. To achieve interactive information fusion, we design a new strategy that learns affine transformation parameters through different convolutions, allowing information exchange between different variates.
First, \( \textbf{H}_{\text{orig}} \) and \( \textbf{H}_{\text{copy}} \) are projected into hidden states using two different 1D convolutional modules \( \alpha \) and \( \beta \), and then transformed into an exponential format for element-wise multiplication:
\begin{equation}
\textbf{H}^1_{\text{orig}} = \textbf{H}_{\text{orig}} \otimes \exp(\beta(\textbf{H}_{\text{copy}})),
\end{equation}
\begin{equation}
\textbf{H}^1_{\text{copy}} = \textbf{H}_{\text{copy}} \otimes \exp(\alpha(\textbf{H}_{\text{orig}})).
\end{equation}
\noindent This can be seen as a scaling transformation on \( \textbf{H}_{\text{orig}} \) and \( \textbf{H}_{\text{copy}} \), which can better handle complex features and non-linear relationships.

Then, the two scaled features \( \textbf{H}^1_{\text{orig}} \) and \( \textbf{H}^1_{\text{copy}} \) are further projected into two other hidden states using two additional 1D convolutional modules \( \mu \) and \( \nu \), and then subtracted or added from \( \textbf{H}^1_{\text{orig}} \) and \( \textbf{H}^1_{\text{copy}} \) to obtain \( \textbf{H}^2_{\text{orig}} \) and \( \textbf{H}^2_{\text{copy}} \), further fusing the information:
\begin{equation}
\textbf{H}^2_{\text{orig}} = \textbf{H}^1_{\text{orig}} \pm \nu(\textbf{H}^1_{\text{copy}}),
\end{equation}
\begin{equation}
\textbf{H}^2_{\text{copy}} = \textbf{H}^1_{\text{copy}} \mp \mu(\textbf{H}^1_{\text{orig}}).
\end{equation}

Finally, \( \textbf{H}^2_{\text{orig}} \) and \( \textbf{H}^2_{\text{copy}} \) are passed through a layer of feature extraction with different kernel sizes. The smaller kernel \( \rho \) captures more detailed local patterns, while the larger kernel \( \omega \) recognizes longer dependencies:
\begin{equation}
\textbf{H}^3_{\text{orig}} = \rho(\textbf{H}^2_{\text{orig}}) \otimes \text{GELU}(\omega(\textbf{H}^2_{\text{copy}})),
\end{equation}
\begin{equation}
\textbf{H}^3_{\text{copy}} = \omega(\textbf{H}^2_{\text{copy}}) \otimes \text{GELU}(\rho(\textbf{H}^2_{\text{orig}})).
\end{equation}
\noindent Subsequently, the features \( \textbf{H}^3_{\text{orig}} \) and \( \textbf{H}^3_{\text{copy}} \) activated by the \textcolor{black}{Gaussian Error Linear Unit (GELU)} activation function are added together, and then processed through a final 1D convolution to obtain \( \textbf{H}_{\text{IFM}} \) for input into the convolutional head:
\begin{equation}
\textbf{H}_{\text{IFM}} = \text{Conv1D}(\textbf{H}^3_{\text{orig}} + \textbf{H}^3_{\text{copy}}).
\end{equation}

\subsection{Training Strategy}

Inspired by PatchTST \cite{nietime}, we introduce a self-supervised pre-training scheme. 
This approach uses a masked autoencoder paradigm to process sequential time series signals, enabling the model to learn high-level representation from unlabeled data, thus expanding the applicability of FusAD across multiple tasks.
Specifically, the entire process is divided into pre-training and formal training phases. 
Algorithm \ref{algorithm2} provides pseudocode for the training phase.

\subsubsection{Pre-training Stage}
During the pre-training stage, an adaptive threshold is used to select masks for the input sequence, and FusAD accurately reconstructs these masked segments. 
This method forces the model to learn useful feature representations during the reconstruction task, inferring the interdependencies within the data. 
We use the Mean Squared Error (MSE) loss function \( \mathcal{L}_{MSE} \) to optimize the model's ability, helping the model understand the intrinsic structure of the data. \textcolor{black}{Furthermore, this self-supervised objective acts as an effective regularization technique, compelling the model to learn generalizable representations and thereby mitigating the risk of overfitting.}
\begin{equation}
\mathcal{L}_{MSE} = \frac{1}{\sum_{i=1}^{T} \lambda_i} \sum_{i=1}^{T} \lambda_i \cdot (x_i - \hat{x}_i)^2,
\end{equation}
\noindent where \( T \) is the length of the sequence, \( x_i \) is the true value of the \( i \)-th time point, \( \hat{x}_i \) is the predicted value of the \( i \)-th time point, and \( \lambda_i \) is the mask value, determining whether to calculate the error for that time point.

\begin{table*}[ht]
\centering
\caption{Classification comparison with state-of-the-art representation learning methods in the case-by-case paradigm.}

\setlength{\tabcolsep}{6pt}
\renewcommand{\arraystretch}{1.1}
\begin{tabular}{ll|cccccccccc}
\toprule
\multicolumn{2}{c}{\textbf{Method}} & \textbf{FusAD} & \textbf{FreRA} & \textbf{TVNet} & \textbf{TS-GAC} & \textbf{Data2Vec} & \textbf{TimesNet} & \textbf{PatchTST} & \textbf{TS2Vec} & \textbf{TS-TCC} & \textbf{TNC} \\
\midrule
\multirow{3}{*}{\shortstack[l]{125 UCR }} 
& Avg. Acc      & \cellcolor{lightgray!30} \textbf{0.863} & 0.850 & 0.844 & 0.831 & 0.832 & 0.823 & 0.825 & 0.818 & 0.780 & 0.776      \\
& Avg. Rank     & \cellcolor{lightgray!30}\textbf{1.956} & 3.128 & 3.985 & 3.887 & 4.798 & 5.850 & 6.474 & 6.565 & 7.228   &7.562\\
& Num. Top-1    & \cellcolor{lightgray!30}\textbf{58}    & 13     & 5     & 2    & 0     & 0     & 0     & 0     &0   &0\\
\midrule
\multirow{3}{*}{\shortstack[l]{30 UEA }} 
& Avg. Acc      & \cellcolor{lightgray!30}\textbf{0.765} & 0.754 & 0.744 & 0.739 & 0.738 & 0.728 & 0.730 & 0.704 & 0.668 & 0.670    \\
& Avg. Rank     & \cellcolor{lightgray!30}\textbf{2.236} & 2.617 & 2.987 & 3.540 & 4.863 & 4.975 & 5.821 & 6.310 & 6.622 & 7.233   \\
& Num. Top-1   & \cellcolor{lightgray!30}\textbf{14}     & 6 &  4     & 1     & 1     & 0     & 0     & 2     & 0     & 0  \\
\bottomrule
\end{tabular}
\label{table1}
\end{table*}

\begin{figure*}[ht]
    \begin{subfigure}[b]{0.48\linewidth}
        
        \includegraphics[width=\linewidth]{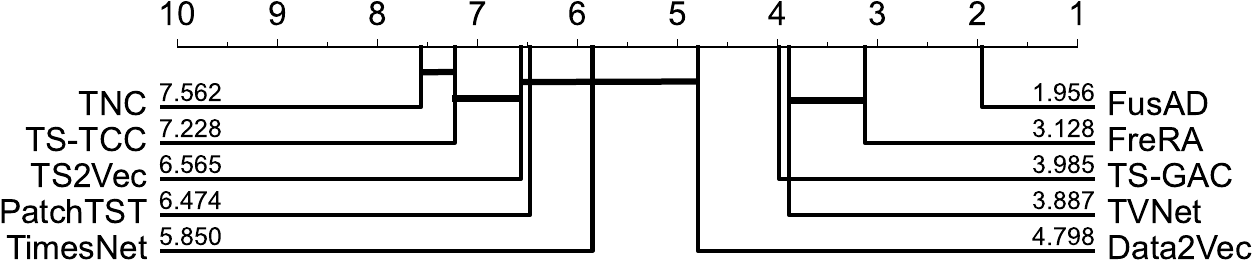}
        \label{fig:sub1}
    \end{subfigure}
    \hfill
    \begin{subfigure}[b]{0.48\linewidth}
        
        \includegraphics[width=\linewidth]{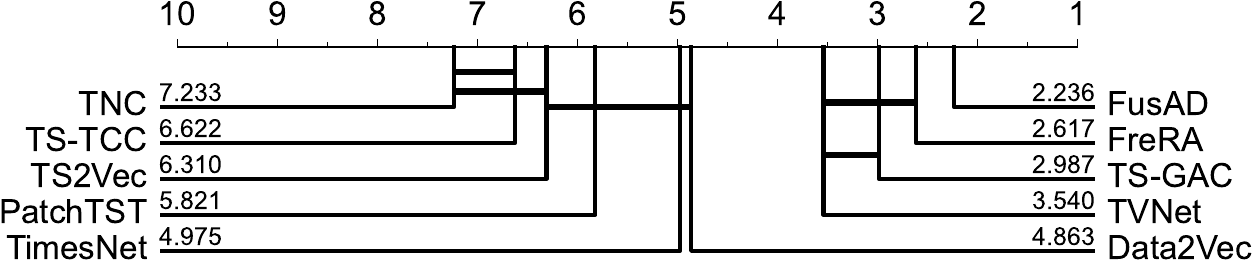}
        \label{fig:sub2}
    \end{subfigure}
    \caption{\textcolor{black}{CD diagram of representation learning methods on UCR (left) and UEA (right) datasets with a confidence level of 95\%.}}
    \label{fig4}
    \vspace{-0.3cm}
\end{figure*}

\subsubsection{Task-specific Training Stage}
After pre-training, we use the optimal weights to initialize the model and conduct supervised training for different tasks:
\begin{itemize}
    \item Classification: we use the label-smooth cross-entropy loss function \( \mathcal{L}_{SCE} \) for optimization. 
By introducing label smoothing, we fine-tune the complete data, which helps improve the model's robustness and enhance its classification performance.
\begin{equation}
\mathcal{L}_{SCE} = -\sum_{i=1}^{k} y_i^{\text{smooth}} \cdot \log(\hat{y}_i),
\end{equation}
\begin{equation}
y_i^{\text{smooth}} = (1 - \epsilon) \cdot y_i + \frac{\epsilon}{k},
\end{equation}
\noindent where \( y_i^{\text{smooth}} \) is the true class label of the one-hot encoded label after label smoothing, \( \hat{y}_i \) is the predicted probability of each class, \( k \) is the number of classes, and label smoothing adjusts the true label's smooth parameter \( \epsilon \) (usually a small value) to prevent the model from overconfidence and enhance its generalization ability.
    \item Forecasting \& Anomaly Detection: For both forecasting and anomaly detection tasks, we use the MSE loss to measure the discrepancy between the predicted and actual values. In particular, for anomaly detection, a large prediction error can be used to identify abnormal patterns.
\end{itemize}


\begin{algorithm}[t]
\caption{FusAD Training Strategy}
\textbf{Input:} Training samples $\mathbf{X}$, label $\mathbf{y}$
\textbf{Output:} Trained model parameters $\Theta$
\begin{algorithmic}[1]
\State \textbf{Pre-training:}
\While{not converged}
    \State Randomly mask a portion of input $\mathbf{X}$
    \State Obtain reconstructed output $\hat{\mathbf{X}} = \text{FusAD}(\mathbf{X}_{\text{masked}})$
    \State Compute loss $L_{\mathrm{MSE}} = \text{MSE}(\hat{\mathbf{X}}, \mathbf{X}_{\text{true}})$
    \State Update parameters $\Theta$ using $L_{\mathrm{MSE}}$
\EndWhile
\State \textbf{Fine-tuning:}
\While{not converged}
    \State Get output $(\hat{\mathbf{y}}^{(c)}, \hat{\mathbf{X}}^{(f)}, \hat{\mathbf{a}}^{(a)}) = \text{FusAD}(\mathbf{X})$
    \If{Classification task}
        \State Compute $L_{\mathrm{SCE}} = \text{LabelSmoothedCE}(\hat{\mathbf{y}}^{(c)}, \mathbf{y})$
    \ElsIf{Forecasting or Anomaly task}
        \State Compute $L_{\mathrm{MSE}} = \text{MSE}(\hat{\mathbf{X}}^{(f)}, \mathbf{X}_{\text{true}})$ or $L_{\mathrm{MSE}} = \text{MSE}(\hat{\mathbf{a}}^{(a)}, \mathbf{a}_{\text{true}})$
    \EndIf
    \State Update parameters $\Theta$ using corresponding loss
\EndWhile
\State \Return $\Theta$
\end{algorithmic}

\label{algorithm2}

\end{algorithm}

\section{Multi-task Experiments}
\label{experiment}
In this section, we evaluate the effectiveness of FusAD in time series classification, prediction, and anomaly detection tasks. 
We demonstrate that our FusAD can serve as a base model with competitive performance on these tasks.

\subsection{Classification}
\subsubsection{Datasets}
We evaluate FusAD on 125 univariate datasets from the UCR archive \cite{dau2019ucr} and 30 multivariate datasets from the UEA archive \cite{bagnall2018uea}. 
The UCR archive is one of the most comprehensive collections for TSA, each dataset presenting unique challenges and characteristics spanning domains such as healthcare, finance, and environmental monitoring. 
The UEA archive comprises a diverse set of real-world multidimensional time series datasets, including human activity recognition, sensor data interpretation, and complex monitoring systems.

\subsubsection{Baselines and Implementation Details}
We compare FusAD with several baseline methods under three paradigms. The case-by-case paradigm includes representation learning methods (e.g., TS-TCC \cite{eldele2021time}, TS2Vec \cite{yue2022ts2vec}, Data2Vec \cite{pieper2023self}), TSA approaches (e.g., PatchTST \cite{nietime}, TimesNet \cite{wutimesnet}, TVNet \cite{litvnet}), and time series classification methods (e.g., TNC \cite{tonekaboniunsupervised}, TS-GAC \cite{wang2024graph}, FreRA \cite{tian2025frera}). 
All competing methods are trained under the case-by-case setting. 
We implement FusAD in PyTorch on a single NVIDIA A800 80GB GPU. 
AdamW is used for optimization, with an initial learning rate of $1\times10^{-3}$, weight decay of $1\times10^{-4}$, and batch size of 128 during pre-training, which are adopted for both pre-training and fine-tuning stages. 
The pre-training stage is run for 100 epochs, while the fine-tuning stage runs for 200 epochs.

\subsubsection{Evaluation Metrics}
Following TS2Vec \cite{yue2022ts2vec}, we employ several widely accepted criteria for classifier evaluation, including the number of datasets achieving the highest accuracy (Num. Top-1), average accuracy (Avg. ACC), average rank (Avg. Rank) \cite{hossin2015review}, and the Critical Difference (CD) diagram \cite{demvsar2006statistical}.

\begin{table*}[h]
\centering
\caption{Multivariate long-term series forecasting results on different prediction lengths $\in \{96, 192, 336, 720\}$. A lower value indicates better performance. The best results are in \textbf{Bold} and the second best ones are \underline{underlined}.}
\begin{center}
\begin{small}
\scalebox{0.75}{
\setlength\tabcolsep{3pt}
\renewcommand{\arraystretch}{1.1}
\begin{tabular}{c|c|cc|cc|cc|cc|cc|cc|cc|cc|cc|cc|cc|cc}
\toprule

\multicolumn{2}{c}{Methods}&\multicolumn{2}{c}{FusAD}&\multicolumn{2}{c}{AMD}&\multicolumn{2}{c}{Affirm}&\multicolumn{2}{c}{iTransformer}&\multicolumn{2}{c}{PatchTST}&\multicolumn{2}{c}{Crossformer}&\multicolumn{2}{c}{FEDformer}&\multicolumn{2}{c}{Autoformer}&\multicolumn{2}{c}{RLinear}&\multicolumn{2}{c}{Dlinear}&\multicolumn{2}{c}{TimesNet}&\multicolumn{2}{c}{GPT4TS} \\

\midrule

\multicolumn{2}{c|}{Metric} & MSE  & MAE & MSE  & MAE & MSE & MAE& MSE & MAE& MSE  & MAE& MSE  & MAE& MSE  & MAE& MSE  & MAE& MSE  & MAE& MSE  & MAE& MSE  & MAE& MSE  & MAE \\
\midrule

\multirow{5}{*}{\rotatebox{90}{$Electricity$}}
~  &  96  &  \cellcolor{lightgray!30}\boldres{0.129} &  \cellcolor{lightgray!30}\boldres{0.222} & \boldres{0.129}   &  0.224 & \boldres{0.129} & 0.223&  0.148  &  0.240  &  0.138  &  0.230  &  0.219  &  0.314  &  0.193  &  0.308  &  0.201  &  0.317  &  0.201  &  0.281  &  0.140  &  0.237  &  0.168  &  0.272  &  0.139  &  0.238     \\
~  &  192  &  \cellcolor{lightgray!30} \secondres{0.147} &  \cellcolor{lightgray!30} \boldres{0.238}   & \secondres{0.147}    &  \boldres{0.238}  & \boldres{0.146} & 0.239 &  0.162  &  0.253  & 0.149 & 0.243 &  0.231  &  0.322  &  0.201  &  0.315  &  0.222  &  0.334  &  0.201  &  0.283  &  0.153  &  0.249  &  0.184  &  0.289  &  0.153  &  0.251     \\
~  &  336  & \cellcolor{lightgray!30}\boldres{0.160}  &  \cellcolor{lightgray!30}\boldres{0.251} & \boldres{0.160} & 0.253  &0.162 & \secondres{0.252}&  0.178  &  0.269  &  0.169  &  0.262  &  0.246  &  0.337  &  0.214  &  0.329  &  0.231  &  0.338  &  0.215  &  0.298  &  0.169  &  0.267  &  0.198  &  0.300  &  0.169  &  0.266    \\
~  &  720  &  \cellcolor{lightgray!30} \boldres{0.190}   & \cellcolor{lightgray!30} \boldres{0.285}  & 0.193  &  \secondres{0.286}  & \secondres{0.191} & 0.288&  0.225  &  0.317  &  0.211  &  0.299  &  0.280  &  0.363  &  0.246  &  0.355  &  0.254  &  0.361  &  0.257  &  0.331  &  {0.203} &  0.301  &  0.220  &  0.320  &  0.206  &  {0.297}    \\ \cmidrule{2-26}
~  &  Avg  & \cellcolor{lightgray!30} \boldres{0.156} &  \cellcolor{lightgray!30} \boldres{0.249} &\secondres{0.157}  &  \secondres{0.250} & \secondres{0.157}& \secondres{0.250} &  0.178  &  0.270  &  0.167  &  0.259  &  0.244  &  0.334  &  0.214  &  0.327  &  0.227  &  0.338  &  0.219  &  0.298  &  0.166  &  0.264  &  0.193  &  0.295  &  0.167  &  0.263     \\

\midrule

\multirow{5}{*}{\rotatebox{90}{$ETTh1$}}
~  &  96  &\cellcolor{lightgray!30} \boldres{0.361} &\cellcolor{lightgray!30} \boldres{0.389} &0.369 & 0.397 &\secondres{0.363} & \secondres{0.392}&  0.386  &  0.405  &  0.382  &  0.401  &  0.423  &  0.448  &  0.376  &  0.419  &  0.449  &  0.459  &  0.386  &  0.395  &  0.375  &  0.399  &  0.384  &  0.402  &  0.376  &  0.397      \\
~  &  192  &\cellcolor{lightgray!30} \boldres{0.399}& \cellcolor{lightgray!30} \boldres{0.410}&\secondres{0.401} & \secondres{0.416}  & 0.408& 0.421&  0.441  &  0.436  &  0.428  &  0.425  &  0.471  &  0.474  &  0.420  &  0.448  &  0.500  &  0.482  &  0.437  &  0.424  &  {0.405} & \secondres{0.416} &  0.436  &  0.429  &  0.416  &  0.418    \\
~  &  336  &\cellcolor{lightgray!30} \secondres{0.421} &\cellcolor{lightgray!30} 0.429 &\boldres{0.418} & \secondres{0.427} &0.424 & \boldres{0.426}&  0.487  &  0.458  &  0.451  &  0.436  &  0.570  &  0.546  &  0.459  &  0.465  &  0.521  &  0.496  &  0.479  &  0.446  &  0.439  &  0.443  &  0.491  &  0.469  &  0.442  &  0.433      \\
~  &  720  & \cellcolor{lightgray!30} \secondres{0.443}& \cellcolor{lightgray!30} \boldres{0.453}& \boldres{0.439} &0.454 &0.450 &\boldres{0.453} &  0.503  &  0.491  &  {0.452} &  0.459  &  0.653  &  0.621  &  0.506  &  0.507  &  0.514  &  0.512  &  0.481  &  0.470  &  0.472  &  0.490  &  0.521  &  0.500  &  0.477  &  {0.456}    \\ \cmidrule{2-26}
~  &  Avg  &\cellcolor{lightgray!30} \boldres{0.406} &\cellcolor{lightgray!30} \boldres{0.420} &\secondres{0.407} & 0.424 & 0.411&\secondres{0.423} &  0.454  &  0.448  &  0.428  &  0.430  &  0.529  &  0.522  &  0.440  &  0.460  &  0.496  &  0.487  &  0.446  &  0.434  &  0.423  &  0.437  &  0.458  &  0.450  &  0.428  &  0.426    \\

\midrule

\multirow{5}{*}{\rotatebox{90}{$ETTh2$}}
~  &  96 &\cellcolor{lightgray!30}\boldres{0.266} & \cellcolor{lightgray!30}\boldres{0.331}& \secondres{0.274} & \secondres{0.337} & 0.276& 0.343&  0.297  &  0.349  &  0.285  &  0.340  &  0.745  &  0.584  &  0.358  &  0.397  &  0.346  &  0.388  &  0.288  &  {0.338} &  0.289  &  0.353  &  0.340  &  0.374  &  0.285  &  0.342    \\
~  &  192  &\cellcolor{lightgray!30}\secondres{0.319} &\cellcolor{lightgray!30}\boldres{0.372} & 0.351 & 0.383 &\boldres{0.316} & \boldres{0.372}&  0.380  &  0.400  &  0.356  &  0.386  &  0.877  &  0.656  &  0.429  &  0.439  &  0.456  &  0.452  &  0.374  &  0.390  &  0.383  &  0.418  &  0.402  &  0.414  &  0.354  &  0.389     \\
~  &  336  & \cellcolor{lightgray!30}0.360&\cellcolor{lightgray!30}\secondres{0.395} & 0.375 &0.411  & \boldres{0.341}& \boldres{0.383}&  0.428  &  0.432  & \secondres{0.350} & \secondres{0.395} &  1.043  &  0.731  &  0.496  &  0.487  &  0.482  &  0.486  &  0.415  &  0.426  &  0.448  &  0.465  &  0.452  &  0.452  &  0.373  &  0.407     \\
~  &  720  &\cellcolor{lightgray!30}\secondres{0.392} &\cellcolor{lightgray!30}\boldres{0.425} &0.402 & 0.438 & \boldres{0.391}& 0.429 &  0.427  &  0.445  &  {0.395} & \secondres{0.427} &  1.104  &  0.763  &  0.463  &  0.474  &  0.515  &  0.511  &  0.420  &  0.440  &  0.605  &  0.551  &  0.462  &  0.468  &  0.406  &  0.441    \\ \cmidrule{2-26}
~  &  Avg  &\cellcolor{lightgray!30}\secondres{0.334} & \cellcolor{lightgray!30}\boldres{0.380} &0.351 &0.392  &\boldres{0.331} & \secondres{0.381} &  0.383  &  0.407  &  0.347  &  0.387  &  0.942  &  0.684  &  0.437  &  0.449  &  0.450  &  0.459  &  0.374  &  0.399  &  0.431  &  0.447  &  0.414  &  0.427  &  0.355  &  0.395     \\

\midrule

\multirow{5}{*}{\rotatebox{90}{$ETTm1$}}
~  &  96  &\cellcolor{lightgray!30} 0.288 & \cellcolor{lightgray!30}\boldres{0.338} &\boldres{0.284} & \secondres{0.339} &\secondres{0.285} &0.344 &  0.334  &  0.368  &  0.291  &  {0.340} &  0.404  &  0.426  &  0.379  &  0.419  &  0.505  &  0.475  &  0.355  &  0.376  &  0.299  &  0.343  &  0.338  &  0.375  &  0.292  &  0.346     \\
~  &  192  &\cellcolor{lightgray!30} 0.326& \cellcolor{lightgray!30} 0.367&\boldres{0.322} & \boldres{0.362} &\secondres{0.323} &\secondres{0.365} &  0.377  &  0.391  &  0.328  & \secondres{0.365} &  0.450  &  0.451  &  0.426  &  0.441  &  0.553  &  0.496  &  0.391  &  0.392  &  0.335  &  \secondres{0.365}&  0.374  &  0.387  &  0.332  &  0.372     \\
~  &  336  &\cellcolor{lightgray!30} \secondres{0.355} & \cellcolor{lightgray!30} \secondres{0.381}&0.360 & \boldres{0.380} &\boldres{0.351} & 0.384&  0.426  &  0.420  &  0.365  &  0.389  &  0.532  &  0.515  &  0.445  &  0.459  &  0.621  &  0.537  &  0.424  &  0.415  &  0.369  & {0.386} &  0.410  &  0.411  &  0.366  &  0.394     \\
~  &  720  &\cellcolor{lightgray!30} \boldres{0.412} &\cellcolor{lightgray!30}\boldres{0.413} & 0.421 & 0.416 &  {0.418}& \secondres{0.414}&  0.491  &  0.459  &  0.422  &  0.423  &  0.666  &  0.589  &  0.543  &  0.490  &  0.671  &  0.561  &  0.487  &  0.450  &  0.425  &  {0.421} &  0.478  &  0.450  & \secondres{0.417} &  0.421     \\ \cmidrule{2-26}
~  &  Avg  &\cellcolor{lightgray!30} \secondres{0.345} &\cellcolor{lightgray!30} \boldres{0.374} & 0.347& \boldres{0.374} & \boldres{0.344}& 0.377&  0.407  &  0.410  &  0.352  &  {0.379} &  0.513  &  0.495  &  0.448  &  0.452  &  0.588  &  0.517  &  0.414  &  0.408  &  0.357  &  0.379  &  0.400  &  0.406  &  0.352  &  0.383   \\

\midrule

\multirow{5}{*}{\rotatebox{90}{$ETTm2$}}
~  &  96  & \cellcolor{lightgray!30} \boldres{0.161}&\cellcolor{lightgray!30} \boldres{0.254} &\secondres{0.167}  &  {0.258} & \secondres{0.167}& 0.260&  0.180  &  0.264  &  0.169  & \boldres{0.254} &  0.287  &  0.366  &  0.203  &  0.287  &  0.255  &  0.339  &  0.182  &  0.265  & \secondres{0.167} &  0.260  &  0.187  &  0.267  &  0.173  &  0.262     \\
~  &  192  &\cellcolor{lightgray!30} \boldres{0.220} &\cellcolor{lightgray!30} \boldres{0.293} & \secondres{0.221} & \secondres{0.294} & \secondres{0.221}&0.296 &  0.250  &  0.309  &  0.230  & \secondres{0.294} &  0.414  &  0.492  &  0.269  &  0.328  &  0.281  &  0.340  &  0.246  &  0.304  &  0.224  &  0.303  &  0.249  &  0.309  &  0.229  &  0.301     \\
~  &  336  &\cellcolor{lightgray!30} 0.272 &\cellcolor{lightgray!30} \boldres{0.326} &\boldres{0.270} &\secondres{0.327}  &\secondres{0.271} &0.328 &  0.311  &  0.348  &  0.280  &  0.329  &  0.597  &  0.542  &  0.325  &  0.366  &  0.339  &  0.372  &  0.307  &  0.342  &  0.281  &  0.342  &  0.321  &  0.351  &  0.286  &  0.341    \\
~  &  720  & \cellcolor{lightgray!30} \boldres{0.349}&\cellcolor{lightgray!30} \secondres{0.379} &0.356 & 0.382 & \secondres{0.351}& \boldres{0.377}&  0.412  &  0.407  &  0.378  &  0.386  &  1.730  &  1.042  &  0.421  &  0.415  &  0.433  &  0.432  &  0.407  &  0.398  &  0.397  &  0.421  &  0.408  &  0.403  &  0.378  &  0.401     \\ \cmidrule{2-26}
~  &  Avg  & \cellcolor{lightgray!30} \boldres{0.250}& \cellcolor{lightgray!30} \boldres{0.313}&0.254 &\secondres{0.315} &\secondres{0.252} & \secondres{0.315}&  0.288  &  0.332  &  0.264  &  0.316  &  0.757  &  0.611  &  0.305  &  0.349  &  0.327  &  0.371  &  0.286  &  0.327  &  0.267  &  0.332  &  0.291  &  0.333  &  0.267  &  0.326     \\

\midrule

\multirow{5}{*}{\rotatebox{90}{$Exchange$}}
~  &  96  & \cellcolor{lightgray!30} \secondres{0.081}& \cellcolor{lightgray!30} \secondres{0.199}& 0.083 & 0.201  & \boldres{0.080}& \boldres{0.198}&  0.086  &  0.206  &  0.088  &  0.205  &  0.256  &  0.367  &  0.148  &  0.278  &  0.197  &  0.323  &  0.093  &  0.217  & \secondres{0.081} &  0.203  &  0.107  &  0.234  &  {0.082} & \secondres{0.199}     \\
~  &  192  &\cellcolor{lightgray!30} \secondres{0.169} &\cellcolor{lightgray!30} \boldres{0.293} &0.171 & \boldres{0.293} & \secondres{0.169}&0.296 &  0.177  &  0.299  &  0.176  &  0.299  &  0.470  &  0.509  &  0.271  &  0.315  &  0.300  &  0.369  &  0.184  &  0.307  & \boldres{0.157} & \boldres{0.293} &  0.226  & 0.344  &  {0.171} & \boldres{0.293}    \\
~  &  336 &\cellcolor{lightgray!30} 0.308 & \cellcolor{lightgray!30} 0.405 &0.309 & \secondres{0.402}  &0.325 &0.411 &  0.331  &  0.417  & \boldres{0.301} & \boldres{0.397} &  1.268  &  0.883  &  0.460  &  0.427  &  0.509  &  0.524  &  0.351  &  0.432  & \secondres{0.305} &  {0.414} &  0.367  &  0.448  &  0.354  &  0.428    \\
~  &  720 & \cellcolor{lightgray!30} \secondres{0.701} &\cellcolor{lightgray!30} \secondres{0.618} &0.750 &  0.652 & 0.852& 0.690&  {0.847} &  {0.691} &  0.901  &  0.714  &  1.767  &  1.068  &  1.195  &  0.695  &  1.447  &  0.941  &  0.886  &  0.714  & \boldres{0.643} & \boldres{0.601} &  0.964  &  0.746  &  0.877  &  0.704    \\ \cmidrule{2-26}
~  &  Avg  &\cellcolor{lightgray!30} \secondres{0.314} &\cellcolor{lightgray!30} \secondres{0.379} &0.328 &  0.387 & 0.356&0.399 &  {0.360} &  {0.403} &  0.367  &  0.404  &  0.940  &  0.707  &  0.519  &  0.429  &  0.613  &  0.539  &  0.379  &  0.418  & \boldres{0.297} & \boldres{0.378} &  0.416  &  0.443  &  0.371  &  0.406   \\

\midrule

\multirow{5}{*}{\rotatebox{90}{$Traffic$}}
~  &  96  &\cellcolor{lightgray!30} \boldres{0.362} & \cellcolor{lightgray!30} \boldres{0.252}& 0.366 & 0.259 &\secondres{0.364} &\secondres{0.255} &  0.395  &  0.268  &  0.401  &  0.267  &  0.522  &  0.290  &  0.587  &  0.366  &  0.613  &  0.388  &  0.649  &  0.389  &  0.410  &  0.282  &  0.593  &  0.321  &  0.388  &  0.282     \\
~  &  192  &\cellcolor{lightgray!30} \boldres{0.381} & \cellcolor{lightgray!30} \boldres{0.262} & \boldres{0.381} &  0.265 &\boldres{0.381} &\boldres{0.262}&  0.417  &  0.276  &  0.406  &  0.268  &  0.530  &  0.293  &  0.604  &  0.373  &  0.616  &  0.382  &  0.601  &  0.366  &  0.423  &  0.287  &  0.617  &  0.336  &  0.407  &  0.290      \\
~  &  336  & \cellcolor{lightgray!30} \secondres{0.395}& \cellcolor{lightgray!30} \boldres{0.268}& 0.397  & 0.269 & \boldres{0.392}& \boldres{0.268}&  0.433  &  0.283  &  0.421  &  0.277  &  0.558  &  0.305  &  0.621  &  0.383  &  0.622  &  0.337  &  0.609  &  0.369  &  0.436  &  0.296  &  0.629  &  0.336  &  0.412  &  0.294    \\
~  &  720  & \cellcolor{lightgray!30} 0.435 & \cellcolor{lightgray!30} \boldres{0.290}&\boldres{0.429} & 0.292 &\secondres{0.433} & \boldres{0.290}&  0.467  &  0.302  &  0.452  &  0.297  &  0.589  &  0.328  &  0.626  &  0.382  &  0.660  &  0.408  &  0.647  &  0.387  &  0.466  &  0.315  &  0.640  &  0.350  &  0.450  &  0.312  \\ \cmidrule{2-26}
~  &  Avg  & \cellcolor{lightgray!30} \secondres{0.393}&\cellcolor{lightgray!30} \boldres{0.268} &\secondres{0.393} & 0.271 &\boldres{0.392} &\boldres{0.268} &  0.428  &  0.282  &  0.420  &  0.277  &  0.550  &  0.304  &  0.610  &  0.376  &  0.628  &  0.379  &  0.627  &  0.378  &  0.434  &  0.295  &  0.620  &  0.336  &  0.414  &  0.295  \\

\midrule

\multirow{5}{*}{\rotatebox{90}{$Weather$}}
~  &  96  &\cellcolor{lightgray!30} \secondres{0.146} &\cellcolor{lightgray!30} 0.198 &\boldres{0.145} & \secondres{0.197} &\secondres{0.146}& \boldres{0.196}&  0.174  &  0.214  &  0.160  &  0.204  &  0.158  &  0.230  &  0.217  &  0.296  &  0.266  &  0.336  &  0.192  &  0.232  &  0.176  &  0.237  &  0.172  &  0.220  &  0.162  &  0.212     \\
~  &  192  &\cellcolor{lightgray!30} \secondres{0.189} &\cellcolor{lightgray!30} \boldres{0.238} &\boldres{0.187} & \boldres{0.238} & 0.192& 0.239&  0.221  &  0.254  &  0.204  &  0.245  &  0.206  &  0.277  &  0.276  &  0.336  &  0.307  &  0.367  &  0.240  &  0.271  &  0.220  &  0.282  &  0.219  &  0.261  &  0.204  &  0.248     \\
~  &  336  & \cellcolor{lightgray!30} \boldres{0.235} &\cellcolor{lightgray!30} \boldres{0.277} &\secondres{0.240} & 0.280 &0.244 & \secondres{0.278}&  0.278  &  0.296  &  0.257  &  0.285  &  0.272  &  0.335  &  0.339  &  0.380  &  0.359  &  0.395  &  0.292  &  0.307  &  0.265  &  0.319  &  0.280  &  0.306  &  {0.254} &  0.286      \\
~  &  720  &\cellcolor{lightgray!30} \boldres{0.313}&\cellcolor{lightgray!30} \boldres{0.326} &\secondres{0.315} &\secondres{0.330}  &0.321 & 0.332&  0.358  &  0.349  &  0.329  &  0.338  &  0.398  &  0.418  &  0.403  &  0.428  &  0.419  &  0.428  &  0.364  &  0.353  &  {0.323} &  0.362  &  0.365  &  0.359  &  0.326  &  0.337     \\ \cmidrule{2-26}
~  &  Avg  &\cellcolor{lightgray!30} \boldres{0.221} &\cellcolor{lightgray!30}  \boldres{0.260} & \secondres{0.222}& \secondres{0.261} & 0.226&\secondres{0.261} &  0.258  &  0.278  &  0.238  &  0.268  &  0.259  &  0.315  &  0.309  &  0.360  &  0.338  &  0.382  &  0.272  &  0.291  &  0.246  &  0.300  &  0.259  &  0.287  &  0.237  &  0.271     \\

\midrule

\multicolumn{2}{c|}{Count}   &\multicolumn{2}{c|}{\cellcolor{lightgray!30} \boldres{48}/\secondres{22}}  &\multicolumn{2}{c|}{\boldres{17}/\secondres{23}}  &\multicolumn{2}{c|}{\boldres{23}/\secondres{25}} &\multicolumn{2}{c|}{\boldres{0}/\secondres{0}} &\multicolumn{2}{c|}{\boldres{3}/\secondres{4}}  &\multicolumn{2}{c|}{\boldres{0}/\secondres{0}}  &\multicolumn{2}{c|}{\boldres{0}/\secondres{0}}  &\multicolumn{2}{c|}{\boldres{0}/\secondres{0}} &\multicolumn{2}{c|}{\boldres{0}/\secondres{0}}  &\multicolumn{2}{c|}{\boldres{6}/\secondres{5}}  &\multicolumn{2}{c|}{\boldres{0}/\secondres{0}}  &\multicolumn{2}{c}{\boldres{1}/\secondres{2}}     \\

\bottomrule
\end{tabular}
}
\end{small}
\end{center}
\label{table2}
\vspace{-0.3cm}
\end{table*}

\subsubsection{Results}
\textcolor{black}{Table~\ref{table1} provides a detailed comparison between FusAD and various mainstream time series representation learning methods. The results show that FusAD consistently outperforms all baseline methods across all evaluation metrics. On the UCR and UEA datasets, FusAD achieves the highest average accuracies of 0.863 and 0.765 respectively, indicating the strong generalization ability and robustness of FusAD in diverse time series scenarios. Furthermore, FusAD achieves the highest number of datasets with the best accuracy (58 and 14 for UCR and UEA, respectively). In addition, we present the CD diagram (Figure \ref{fig4}) covering all datasets. As illustrated in the diagram, FusAD achieves the best average rank. More importantly, FusAD is not connected by a line to the other competing methods, which demonstrates that its superior classification performance is statistically significant and highlights its robustness and generalization capability across diverse time series domains. }

\begin{table*}[ht]
    \centering
    \caption{The overall anomaly detection results The precision (P), recall (R), and F1-score (F1) values are reported. 
    The best results are in \textbf{Bold} and the second best ones are \underline{underlined}.}
    \setlength{\tabcolsep}{3pt}
    \renewcommand{\arraystretch}{1.1}
    \resizebox{0.999\linewidth}{!}{
    \begin{tabular}{c|ccc|ccc|ccc|ccc|ccc|ccc}
        \toprule
        \multirow{2}{*}{Models}  & \multicolumn{3}{c}{SMD} & \multicolumn{3}{c}{MSL} & \multicolumn{3}{c}{SMAP} & \multicolumn{3}{c}{SWaT} & \multicolumn{3}{c}{PSM} & \multicolumn{3}{c}{Avg} \\
        \cmidrule(lr){2-4} \cmidrule(lr){5-7} \cmidrule(lr){8-10} \cmidrule(lr){11-13} \cmidrule(lr){14-16} \cmidrule(lr){17-19} 
           & P & R & F1 & P & R & F1 & P & R & F1 & P & R & F1 & P & R & F1 & P & R & F1 \\
        \midrule
        \multicolumn{1}{c|}{Informer}  & 0.866 & 0.772 & 0.816 & 0.817 & 0.864 & 0.840 & 0.901 & 0.571 & 0.699 & 0.702 & 0.965 & 0.814 & 0.642 & 0.963 & 0.771 &0.786 & 0.827& 0.788 \\

        \multicolumn{1}{c|}{Autoformer}  & 0.880 & 0.823 & 0.851 & 0.772 & 0.809 & 0.790 & 0.904 & 0.586 & 0.711 & 0.898 & 0.958 & 0.927 & 0.990 & 0.881 & 0.932 & 0.889 & 0.811 & 0.842 \\

       \multicolumn{1}{c|}{Pyraformer}  & 0.856 & 0.806 & 0.830 & 0.838 & 0.859 & 0.848 & 0.925 & 0.577 & 0.710 & 0.879 & 0.960 & 0.917 & 0.716 & 0.960 & 0.820 & 0.843 & 0.832 & 0.825 \\

\multicolumn{1}{c|}{AnomalyTrans}   & \secondres{0.889} & 0.822 & 0.854 & 0.796 & 0.873 & 0.833 & 0.918 & 0.581 & 0.711 & 0.725 & \boldres{0.973} & 0.831 & 0.683 & 0.947 & 0.794 & 0.802 & 0.839 & 0.805 \\

\multicolumn{1}{c|}{Stationary}  & 0.883 & 0.812 & 0.846 & 0.685 & \secondres{0.891} & 0.775 & 0.893 & 0.590 & 0.710 & 0.680 & 0.967 & 0.798 & 0.978 & 0.967 & 0.972 & 0.824 & \secondres{0.845} & 0.820 \\

\multicolumn{1}{c|}{Crossformer}   & 0.830 & 0.766 & 0.797 & \secondres{0.846} & 0.837 & 0.841 & 0.920 & 0.553 & 0.691 & 0.884 & 0.934 & 0.909 & 0.971 & 0.897 & 0.933 & 0.890 & 0.797 & 0.834 \\

\multicolumn{1}{c|}{FEDformer}   & 0.879 & 0.823 & 0.850 & 0.771 & 0.800 & 0.785 & 0.904 & 0.581 & 0.707 & 0.901 & 0.964 & 0.931 & 0.973 & \secondres{0.971} & 0.972 & 0.886 & 0.828 & 0.849 \\

\multicolumn{1}{c|}{DLinear}   & 0.836 & 0.715 & 0.771 & 0.843 & 0.845 & 0.848 & 0.923 & 0.554 & 0.692 & 0.809 & 0.953 & 0.875 & 0.982 & 0.892 & 0.935 & 0.878 & 0.792 & 0.824 \\

\multicolumn{1}{c|}{TimesNet}   & 0.886 & 0.831 & 0.858 & 0.839 & 0.863 & \secondres{0.851} & 0.925 & 0.582 & 0.715 & 0.867 & \boldres{0.973} & 0.917 & 0.981 & 0.967 & 0.974 & 0.900 & 0.843 & 0.863 \\

\multicolumn{1}{c|}{PatchTST}  & 0.874 & 0.816 & 0.844 & 0.840 & 0.862 & 0.851 & 0.924 & 0.575 & 0.709 & 0.807 & 0.944 & 0.872 & \secondres{0.988} & 0.939 & 0.963 & 0.886 & 0.827 & 0.848 \\

\multicolumn{1}{c|}{ModernTCN}  & 0.878 & 0.838 & 0.858 & 0.839 & 0.859 & 0.849 & \boldres{0.951} & 0.576 & 0.712 & 0.918 & 0.959 & 0.938 & 0.980 & 0.963 & 0.972 & \secondres{0.913} & 0.839 & 0.866 \\

\multicolumn{1}{c|}{GPT4TS}  & \boldres{0.898} & \secondres{0.849} & \secondres{0.868} & 0.820 & 0.829 & 0.824 & 0.906 & \secondres{0.609} & 0.728 & \secondres{0.922} & 0.963 & \boldres{0.942} & 0.986 & 0.956 & 0.971 & 0.906 & 0.841 & 0.867 \\

\multicolumn{1}{c|}{TSINR}  & 0.830 & 0.804 & 0.817 & 0.835 & 0.854 & 0.844 & 0.916 & 0.764 & \boldres{0.833} & \boldres{0.993} & 0.723 & 0.936 & \boldres{0.992} & 0.893 & 0.940 & \secondres{0.913} & 0.807 & 0.840 \\

\multicolumn{1}{c|}{TVNet}    & 0.880 & 0.834 & 0.857 & 0.839 & 0.864 & \boldres{0.851} & \secondres{0.929} & 0.582 & 0.716 & 0.912 & 0.963 & 0.937 & 0.983 & 0.967 & \secondres{0.975} & 0.908 & 0.842 & \secondres{0.867} \\

\rowcolor{lightgray!30}
\multicolumn{1}{c|}{FusAD}    & 0.886 & \boldres{0.863} & \boldres{0.879} & \boldres{0.875} & \boldres{0.901} & 0.822 & 0.929 & \boldres{0.625} & \secondres{0.743} & 0.920 & 0.956 & \secondres{0.938} & 0.987 & \boldres{0.980} & \boldres{0.977} & \boldres{0.919} & \boldres{0.865} & \boldres{0.872} \\

        \bottomrule
    \end{tabular}}
    \label{table3}
\end{table*}

\subsection{Forecasting}

\subsubsection{Datasets and Evaluation Metrics}
To assess the effectiveness of FusAD in forecasting tasks, we conduct a comprehensive evaluation on eight benchmark datasets. 
Specifically, these include Electricity\footnote{https://archive.ics.uci.edu/ml/datasets/ElectricityLoadDiagrams20112014}, which contains power consumption data; 
four ETT datasets (ETTh1, ETTh2, ETTm1, ETTm2)\footnote{https://github.com/zhouhaoyi/ETDataset} covering various scenarios of energy transmission technologies;
Exchange \cite{lai2018modeling}, which reflects exchange rate fluctuations; 
Traffic \footnote{https://pems.dot.ca.gov}, which contains traffic flow information; 
and Weather\footnote{https://www.bgc-jena.mpg.de/wetter}, which provides time series of various meteorological variables. 
The Mean Squared Error (MSE) and Mean Absolute Error (MAE) are adopted as the evaluation metrics for multivariate time series forecasting.

\subsubsection{Baselines and Implementation Details}
We compare the performance of FusAD with various state-of-the-art baselines. 
For Transformer-based models, we include comparisons with iTransformer \cite{liuitransformer}, PatchTST \cite{nietime}, Crossformer \cite{zhang2023crossformer}, FEDformer \cite{zhou2022fedformer}, and Autoformer \cite{wu2021autoformer}. 
For MLP and convolution-based models, we compare against RLinear \cite{li2023revisiting}, DLinear \cite{zeng2023transformers}, Affirm \cite{wu2025affirm}, and AMD \cite{hu2025adaptive}. 
For universal time series models, we consider TimesNet \cite{wutimesnet} and GPT4TS \cite{zhou2023one}.
Following the setup of iTransformer \cite{liuitransformer}, we set the input look-back window to 336 time steps for the ETT datasets, 96 for Exchange, 512 for Traffic and Weather, and 96 for Electricity. All datasets are normalized during training. 
For baselines, we report their best results if the experimental setups match; otherwise, we rerun their official codes. 
We set the initial learning rate to 1e-4, weight decay to 1e-6, and train for 20 epochs in the pre-training stage and 50 epochs in the fine-tuning stage.

\subsubsection{Results}

In Table~\ref{table2}, we compare FusAD with various baselines. FusAD consistently outperforms the strong baseline models across all eight datasets, demonstrating powerful predictive capability for long-horizon multivariate time series. Specifically, out of 80 metrics across different scenarios, FusAD ranks within the top two in 70 cases, with 48 first-place results. This significantly exceeds other baseline methods, including the current state-of-the-art Transformer-based model, iTransformer.
FusAD exhibits particularly outstanding performance on the ETT and Weather datasets, achieving consistently better results than other SOTA methods across almost all metrics. On larger-scale datasets such as Electricity and Traffic (with more than 300 and 800 channels, respectively), the performance of FusAD declines slightly, which may be attributed to its relatively limited model capacity. Nevertheless, even in these challenging scenarios, FusAD remains competitive and often outperforms other models.
We attribute this to the spectral module's ability to effectively handle noise in the data, which enhances the model's robustness, and the information interaction module's capacity to better capture both the continuity and global characteristics among multiple variables. Thus, FusAD proves effective when dealing with datasets of varying characteristics and complexities. In particular, for datasets with higher volatility (such as ETTh1 and Weather), we are more inclined to adopt FusAD as the forecasting model.

\subsection{Anomaly Detection}

\subsubsection{Datasets and Evaluation Metrics}
To evaluate the effectiveness of FusAD in anomaly detection, we conduct assessments on five widely used datasets: SMD \cite{su2019robust} for server monitoring, MSL \cite{hundman2018detecting} for space telemetry, SMAP \cite{hundman2018detecting} for Earth observation, SWaT \cite{mathur2016swat} for water treatment security, and PSM \cite{abdulaal2021practical} for industrial pump sensors. Each dataset represents a distinct application domain. 
Following TSINR \cite{li2024tsinr}, we use precision (P), recall (R), and F1-score as evaluation metrics, which are commonly employed in unsupervised time series anomaly detection.

\subsubsection{Baselines and Implementation Details}
We compare the performance of FusAD with 14 state-of-the-art baselines. Among them are several Transformer-based methods, such as Informer \cite{zhou2021informer}, Autoformer \cite{wu2021autoformer}, Prayformer \cite{liupyraformer}, AnomalyTrans \cite{xuanomaly}, Stationary \cite{liu2022non}, Crossformer \cite{zhang2023crossformer}, FEDformer \cite{zhou2022fedformer}, and PatchTST \cite{nietime}. For MLP and convolutional models, we include DLinear \cite{zeng2023transformers}, ModernTCN \cite{luo2024moderntcn}, and TSINR \cite{li2024tsinr} in our comparisons. 
For more general time series models, we compare against TimesNet \cite{wutimesnet}, GPT4TS \cite{zhou2023one}, and TVNet \cite{litvnet}. We follow the same experimental setup as used in GPT4TS. 
For data preparation, each dataset is split using a sliding window. We set the initial learning rate to 1e-4, weight decay to 1e-6, with 20 epochs for the pre-training stage and 50 epochs for the fine-tuning stage.

\subsubsection{Results}
Table~\ref{table3} presents the results, where FusAD achieves the best performance on most datasets, attaining an overall F1-score of 87.21\%. It scores 87.92\% and 97.73\% on the SMD and PSM datasets, notably outperforming advanced models such as TVNet and ModernTCN. TVNet follows closely, with an overall average of 86.76\%. 
Its robust capabilities make it highly effective in anomaly detection, although it falls slightly behind. 
Notably, Transformer-based models tend to underperform in anomaly detection tasks, possibly because the attention mechanism focuses on dominant normal points and thus misses rare anomalies. 
Models that consider periodicity, such as TimesNet and FEDformer, perform well, indicating the importance of periodic analysis in highlighting unusual patterns.

\section{Model Analysis}
\label{model analysis}
\subsection{Ablation Study}

\begin{table}[ht]
    \centering
    \caption{Study the impact of each component of the system. ``Cls., Fore., Ano." respectively represent classification, prediction and anomaly detection tasks, and the evaluation metrics used are accuracy, MSE and F1, respectively. AWR and SRSCP are respectively the Articulatory Word Recognition and SelfRegulationSCP1 in the UEA dataset.
    }
    \resizebox{0.99\linewidth}{!}{
    \begin{tabular}{l|cc|cc|cc}
        \toprule
        \textbf{Variant} 
        & \multicolumn{2}{c|}{\textbf{Cls. (Acc)}} 
        & \multicolumn{2}{c|}{\textbf{Fore. (MSE)}} 
        & \multicolumn{2}{c}{\textbf{Ano. (F1)}} \\
        & AWR & SRSCP & ETTh1 & Exchange & SMD  & PSM  \\
        \midrule
w/o ASM     & 0.986 & 0.787 & 0.422 & 0.382 & 0.858 & 0.953\\
w/o ASM-F   & 0.987 & 0.802 & 0.420 & 0.378 & 0.854 & 0.965\\
w/o ASM-T   & 0.988 & 0.814 & 0.409 & 0.376 & 0.868 & 0.970\\
w/o ASM-W   & 0.989 & 0.805 & 0.416 & 0.375 & 0.871 & 0.963\\
w/o IFM     & 0.976 & 0.835 & 0.419 & 0.378 & 0.872 & 0.971\\
w/o Pretrain& 0.979 & 0.876 & 0.408 & 0.373 & 0.870 & 0.975\\
\midrule
FusAD       & 0.990 & 0.921 & 0.403 & 0.361 & 0.879 & 0.977 \\
        \bottomrule
    \end{tabular}
    }
    \label{table4}
\end{table}

Table~\ref{table4} presents the results of the ablation study, which investigates the effect of removing each key component from FusAD across classification, forecasting, and anomaly detection tasks. 
Across all tasks and datasets, we observe that the removal of the ASM (w/o ASM) causes the most significant performance degradation. Specifically, classification accuracy on AWR and SRSCP drops to 0.986 and 0.787, respectively, and the forecasting MSE on ETTh1 and Exchange increases to 0.422 and 0.382. 
The F1-scores for anomaly detection on SMD and PSM also decrease substantially to 0.858 and 0.953. This clearly demonstrates the pivotal role of the ASM in extracting robust spectral features and mitigating noise for general time series modeling.

Further, we analyze the contributions of individual ASM sub-modules. Removing the Fourier transform branch only (w/o ASM-F) leads to a decrease in performance in all tasks, though the impact is less than removing the whole ASM block. 
This is particularly evident in forecasting (ETTh1), where global periodic components modeled by the Fourier transform are essential. Removing the adaptive thresholding mechanism (w/o ASM-T) results in marked performance drops, especially in noisy or variable environments, confirming the importance of adaptive denoising (e.g., SMD F1 drops to 0.868). 
Excluding the wavelet branch (w/o ASM-W) also causes notable reductions, indicating the value of local and multi-scale temporal features, especially on complex datasets like SRSPC and PSM.
\textcolor{black}{Removing IFM (w/o IFM) shows a moderate yet persistent impact across all tasks, suggesting that IFM significantly outperforms traditional fusion methods in modeling complex nonlinear interactions and further enhances the robustness of the learned representations.}
The removal of pretraining (w/o Pretrain) also degrades performance, yet to a lesser degree, highlighting its benefit for generalization and better initialization, especially on the SRSPC and PSM.

\textcolor{black}{Notably, on the SMD anomaly detection dataset, removing the entire ASM module (w/o ASM) results in a slightly higher F1-score than removing only the Fourier transform branch (w/o ASM-F). We attribute this to the synergistic design of the ASM, which relies on the balance between global (Fourier) and local (Wavelet) feature extraction. By removing only the Fourier component, an imbalanced spectral analysis is created. Without the global context to regularize the features, the model may become overly sensitive to the local, high-frequency noise inherent in the SMD dataset.}

\subsection{\textcolor{black}{Efficiency of Adaptive Filter in Noise Reduction}}

\begin{figure}[t]
    \includegraphics[width=0.49\textwidth]{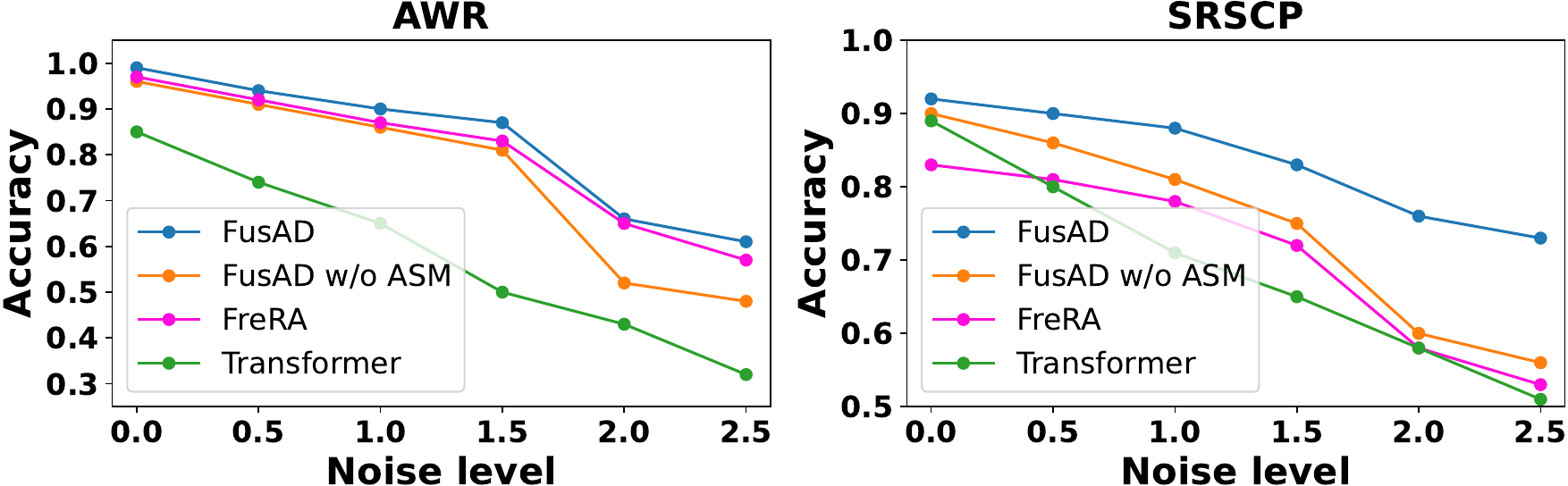}
    \caption{
    \textcolor{black}{Robustness to noise levels on the AWR and SRSCP datasets.}
    }
    \label{fig_noise}
\end{figure}

\textcolor{black}{To conduct an in-depth study of the effectiveness of the adaptive filters in mitigating noise, we compare the robustness of FusAD against representative models from different architectural families: the standard Transformer (attention-based) and FreRA (CNN-based). Specifically, Figure \ref{fig_noise} illustrates the performance of FusAD, with and without ASM, after adding varying levels of Gaussian noise to the time series data from the AWR and SRSCP datasets. As the noise level increases, the performance of both the Transformer and FreRA models degrades significantly. In contrast, FusAD maintains a much more stable and higher accuracy, clearly demonstrating the best noise resistance. The complete FusAD model, equipped with the adaptive filter, shows a remarkably smaller decline in accuracy, underscoring the critical contribution of our adaptive denoising mechanism in complex, noisy environments.}

\begin{figure}[t]
    \includegraphics[width=0.49\textwidth]{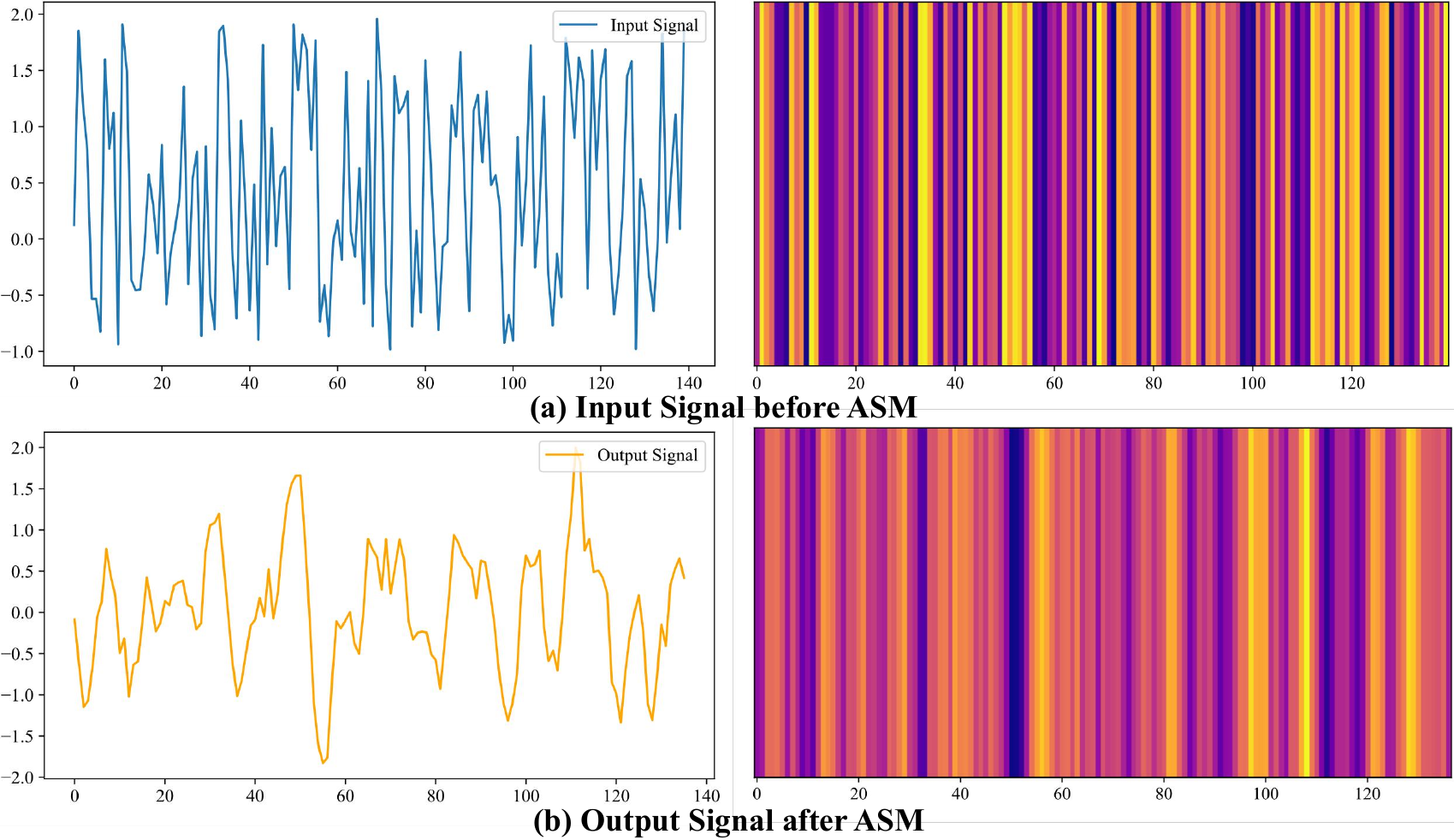}
    \caption{
    The features before (a) and after (b) ASM on the SMD dataset.
    }
    \label{fig_noise_contrast}
    \vspace{-0.3cm}
\end{figure}

As shown in Figure \ref{fig_noise_contrast}, we present the spectral performance and feature distribution of the subsequence on the SMD dataset before and after applying ASM. 
The input sequence exhibits complex high-frequency fluctuations and significant noise interference, with frequent spikes occurring at different time intervals, making the sequence difficult to interpret. 
The feature distribution map displays frequent color changes and lacks clear structure and patterns, indicating that it is challenging to extract features from the raw data.
After applying ASM, the sequence demonstrates a smoother waveform, with a significant reduction in the amplitude of noise spikes in the high-frequency range and more prominent low-frequency components, indicating a remarkable noise suppression effect. 
The feature map shows a more consistent and regular color distribution, suggesting that the features within the sequence have been effectively extracted and enhanced.

\subsection{Scaling Efficiency}
\begin{figure}[ht]
    \begin{subfigure}[b]{0.49\linewidth}
        
        \includegraphics[width=\linewidth]{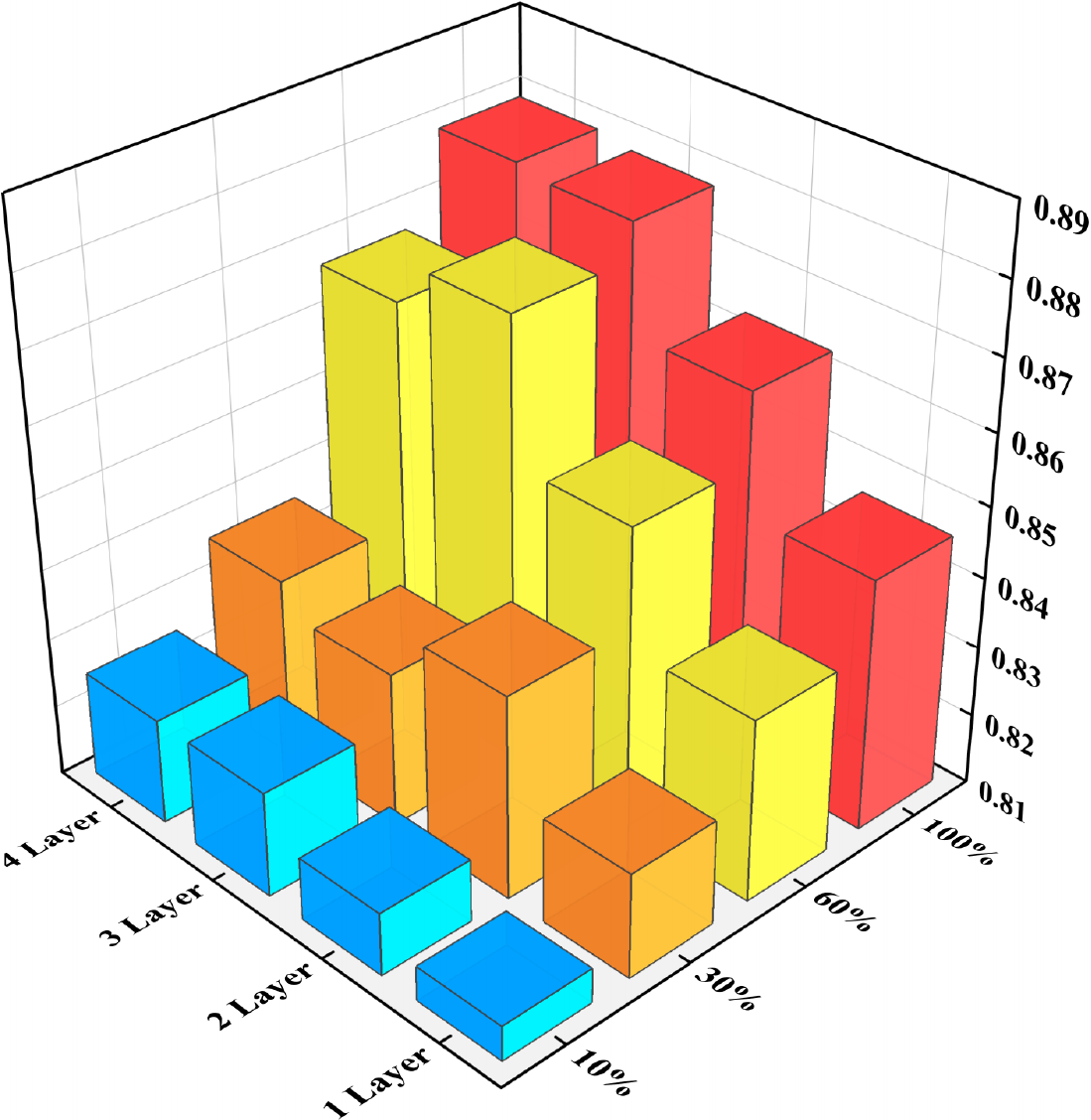}
        \caption{}
    \end{subfigure}
    \hfill
    \begin{subfigure}[b]{0.49\linewidth}
        
        \includegraphics[width=\linewidth]{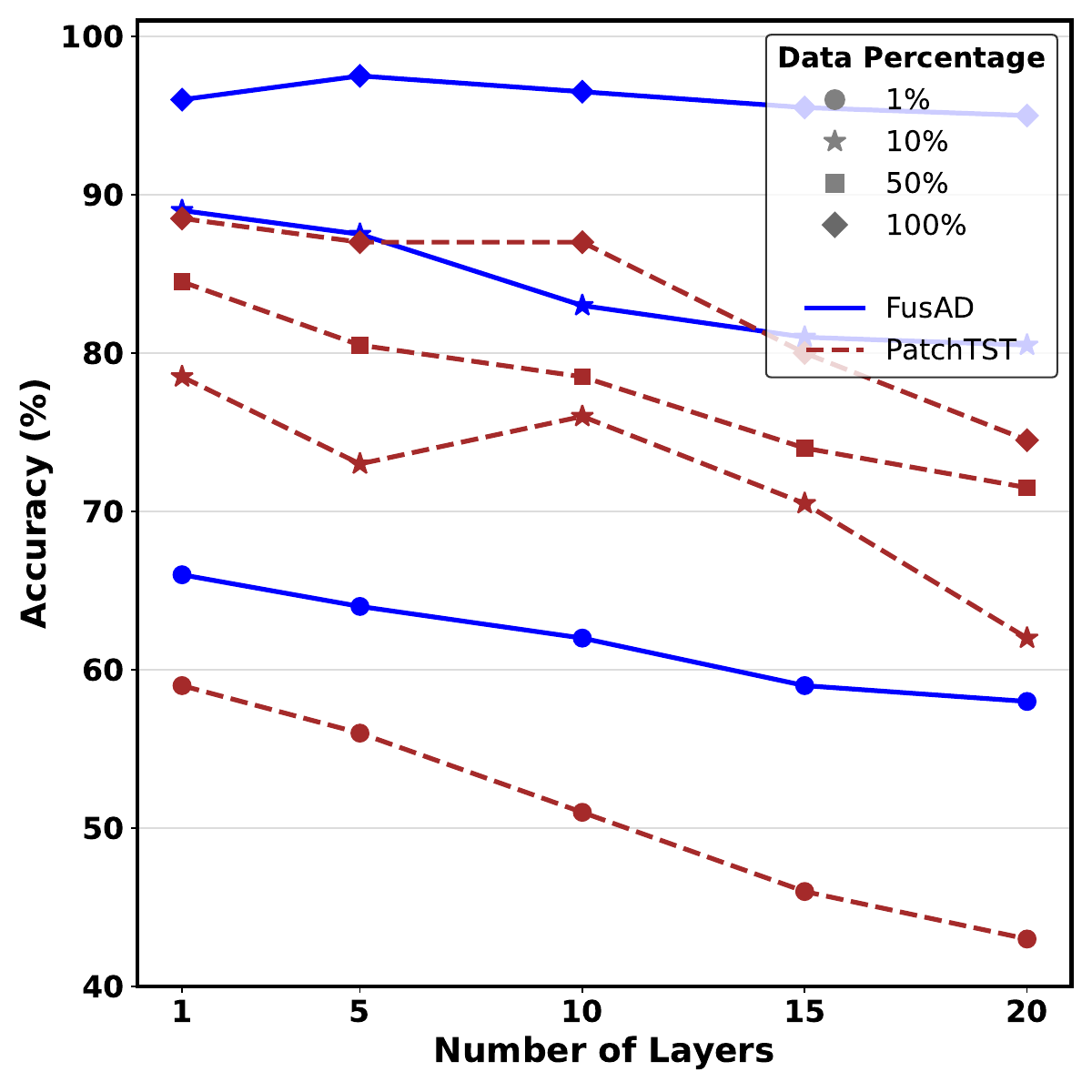}
        \caption{}
    \end{subfigure}
    \caption{\textcolor{black}{Scalability analysis. (a) F1-score of FusAD on SMD with varying percentages of training data and number of layers. (b) Comparison of classification accuracy between FusAD and PatchTST on the uWaveGestureLibraryAll dataset (UCR archive) with varying data percentages and model layers.}}
    \label{fig7}
    \vspace{-0.5cm}
\end{figure}

\textcolor{black}{We compare the scalability of FusAD by observing its performance under different data scales and model depths. First, we conduct experiments using various data scales from the SMD anomaly detection dataset, as shown in Figure \ref{fig7}. At smaller data scales (e.g., 10\% and 30\%), FusAD maintains high F1 scores of 0.825 and 0.836, respectively. As the dataset size increases, the F1 score also improves, reaching 0.879 at the 100\% data scale. This demonstrates that FusAD can effectively utilize more data samples to enhance its performance. With respect to the number of layers, when the layer count increases from 1 to 4, FusAD's performance remains stable or improves slightly, showcasing its robustness even with varying model complexity on this task.}

\textcolor{black}{To further validate the scalability, we also conduct an experiment comparing FusAD with PatchTST on the uWaveGestureLibraryAll dataset. The results show that FusAD maintains remarkably stable and high accuracy even as the model depth increases, whereas PatchTST's performance degrades with more layers, particularly in low-data regimes. This suggests that FusAD is less prone to overfitting and scales more effectively with model complexity. Furthermore, FusAD consistently outperforms PatchTST when trained on smaller subsets of data (e.g., 10\% or 50\%), highlighting its superior data efficiency and ability to learn generalizable representations from limited samples. This validates the adaptability and superior scalability of FusAD in terms of both data volume and model depth.}

\subsection{Training Speed and Memory}

Considering that the hyperparameters and look-back window settings during different training processes may affect the number of model parameters and running speed, we choose three representative models, PatchTST (Transformer-based) \cite{nietime}, DLinear (MLP-based) \cite{zeng2023transformers}, and ModernTCN (CNN-based) \cite{luo2024moderntcn}, and unify their training hyperparameters to compare their parameter counts and training speeds more fairly with the model proposed in this paper.

\begin{figure}[t]
    \includegraphics[width=0.49\textwidth]{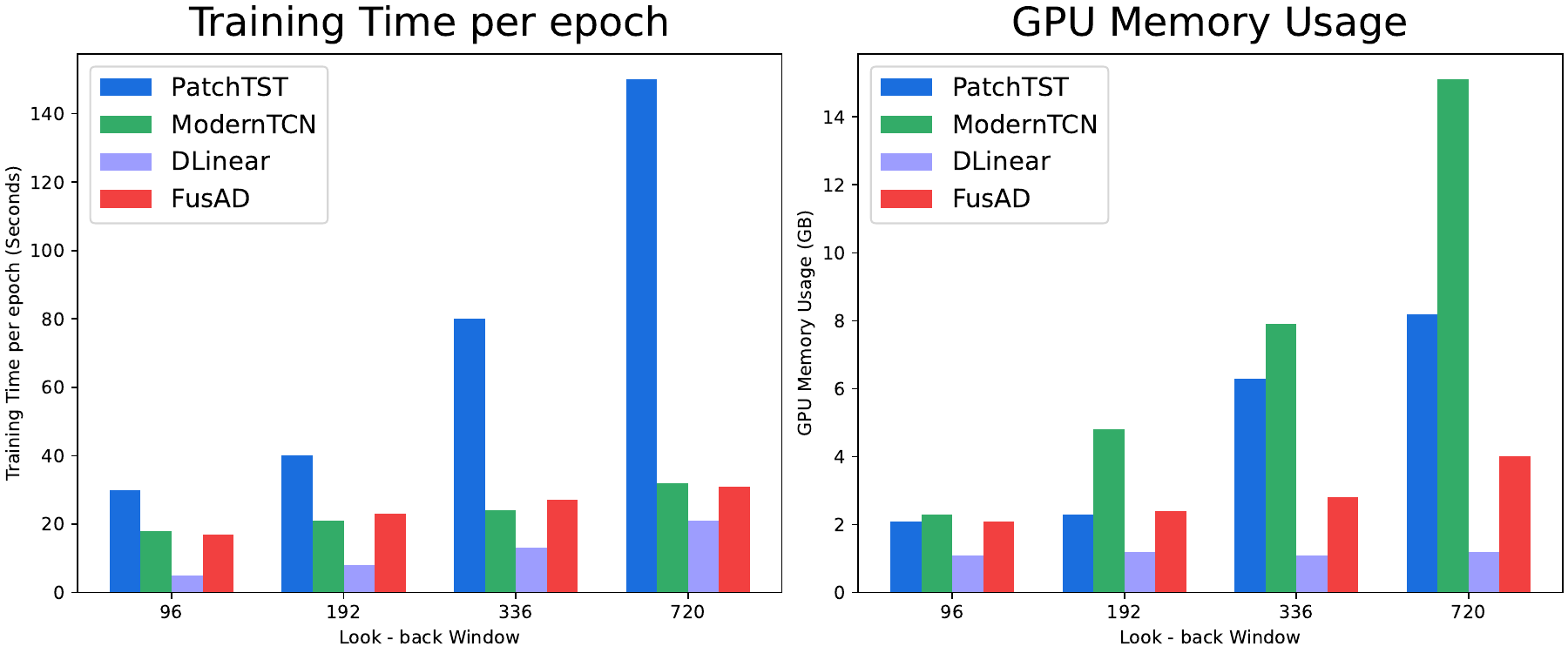}
    \caption{
    Parameters of different input lengths and GPU memory on ETTm2 dataset.
    }
    \label{fig_gpu}
    \vspace{-0.5cm}
\end{figure}

The results on ETTm2 with different input lengths (with the prediction length fixed at 96) are shown in Figure \ref{fig_gpu}. 
As observed in the figure, for PatchTST, the training time increases with the input length, while for ModernTCN, memory usage increases significantly. In contrast, the proposed FusAD model exhibits only a slow increase in training time and memory usage as the input length increases, demonstrating the superiority of the proposed model in terms of efficiency and effectiveness.

\subsection{Sensitivity Analysis}

\subsubsection{Patch Length}

\begin{figure*}[htbp]
    \centering
    \begin{subfigure}{0.99\linewidth}
        \includegraphics[width=\linewidth]{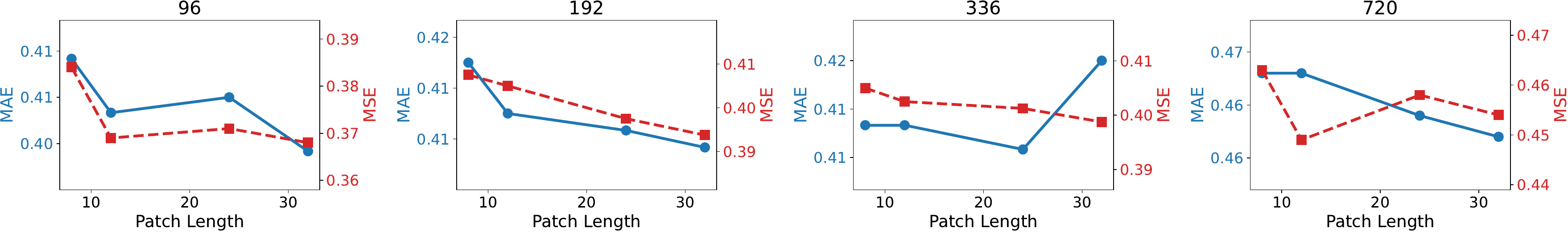}
    \end{subfigure}
    \\
    \begin{subfigure}{0.99\linewidth}
        \includegraphics[width=\linewidth]{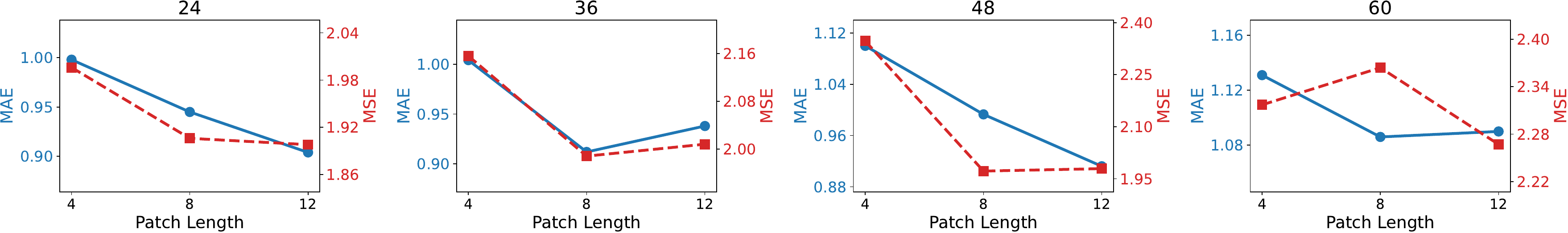}
    \end{subfigure}
    \caption{Comparison of different patch lengths in long-term and short-term forecasting tasks. 
    The top row shows the results on the EETh1 dataset with patch lengths set to 8, 12, 24, and 32 under different prediction lengths (96, 192, 336, and 720). 
    The bottom row presents results on the ILI dataset with patch lengths of 4, 8, and 12 for prediction lengths of 24, 36, 48, and 60. 
    The figure illustrates the trends of MAE and MSE under various parameter settings.}
    \label{fig_patch}
    \vspace{-0.35cm}
\end{figure*}

We evaluate the performance of different patch lengths on both long-term (EETh1) and short-term (ILI \cite{wutimesnet}) forecasting tasks (Figure \ref{fig_patch}). 
The experimental results indicate that varying the patch length does not have a significant impact on the results overall, but FusAD exhibits more robust performance when a moderate patch length is chosen. 
Specifically, using either very small or very large patch lengths tends to degrade forecasting performance. 
For long-term forecasting tasks, we recommend setting the patch length to 24, as it achieves the best balance between accuracy and generalization. 
For short-term forecasting, a patch length of 8 is preferred.

\begin{table}[ht]
\centering
\caption{
The impact of different kernel sizes. A lower mean square error (MSE) or mean absolute error (MAE) indicates better performance. The best results are in \textbf{Bold} and the second best ones are \underline{underlined}.}
\renewcommand\arraystretch{1.1}
\scalebox{0.8}{
\begin{tabular}{ll|cccc|cccc}
\toprule
\multirow{2}{*}{Models} & \multirow{2}{*}{Metrics} 
    & \multicolumn{4}{c|}{ETTh1} 
    & \multicolumn{4}{c}{ETTm1}  \\
    \cmidrule(lr){3-6}     \cmidrule(lr){7-10} 
& & 96 & 192 & 336 & 720 & 96 & 192 & 336 & 720  \\
\midrule
\multirow{2}{*}{$k=1$} 
& MSE & 0.375 & 0.407 & 0.425 & 0.464 & 0.293 & 0.327 & 0.371 & \boldres{0.412}  \\
& MAE & 0.413 & 0.417 & 0.435 & 0.464 & 0.358 & 0.371 & 0.392 & 0.419  \\

\midrule
\multirow{2}{*}{$k=3$} 
& MSE & \boldres{0.361} & \secondres{0.399} & \boldres{0.421} & \boldres{0.443} & \boldres{0.288} & 0.326 & \boldres{0.355} & \boldres{0.412} \\
& MAE & \boldres{0.389} & \secondres{0.410} & 0.429 & \boldres{0.453} & \boldres{0.338} & 0.367 & \boldres{0.381} & \boldres{0.413} \\

\midrule
\multirow{2}{*}{$k=5$} 
& MSE & \secondres{0.369} & \boldres{0.398} & \boldres{0.421} & \secondres{0.444} & 0.289 & \boldres{0.324} & \secondres{0.367} & 0.417  \\
& MAE & \secondres{0.395} & \boldres{0.409} & 0.438 & \secondres{0.457} & 0.339 &\secondres{0.362} & \secondres{0.390} & \secondres{0.415}  \\

\midrule
\multirow{2}{*}{$k=7$} 
& MSE & 0.370 & \secondres{0.399} & 0.428 & 0.450 & 0.291 & \secondres{0.325} & 0.368 & 0.420  \\
& MAE & 0.406 & 0.412 & 0.439 & 0.462 & 0.341 &\boldres{0.360} & 0.391 & 0.426 \\

\bottomrule
\end{tabular}
}
\label{table_kernel}
\end{table}

\subsubsection{Kernel Size}

\textcolor{black}{We also investigated the influence of different convolution kernel sizes during the information interaction stage. 
As shown in Table \ref{table_kernel}, both excessively small and excessively large kernel sizes negatively affect prediction accuracy. 
This study recommends using a 3×3 convolution kernel, as this configuration offers a good balance between capturing local features and maintaining computational efficiency.}

\subsubsection{Mask Ratio}

\begin{table}[ht]
\centering
\caption{Sensitivity experiments of mask ratio (forecasting length is 96). The best results are in \textbf{Bold} and the second best ones are \underline{underlined}.}
\renewcommand\arraystretch{1.1}
\scalebox{0.975}{
\begin{tabular}{c|l|cccccc}
\hline
\multicolumn{2}{c|}{Mask Ratio}  & 0.1 & 0.15 & 0.2 & 0.25 & 0.3 & 0.35\\
\hline
\multirow{2}{*}{Electricity}   
& MSE & 0.148 & 0.139 & \secondres{0.131}    &  \boldres{0.129} & 0.133 & 0.137  \\
& MAE & 0.240 & 0.208 & 0.228    &  \boldres{0.222} & \secondres{0.225} & 0.231  \\
\midrule

\multirow{2}{*}{Exchange} 
& MSE & 0.088 & 0.086 & 0.082    & \boldres{0.081} & \boldres{0.081} & 0.082  \\
& MAE & 0.206 & 0.205 & 0.201    & \secondres{0.199} & \boldres{0.198} & \secondres{0.199}  \\
\midrule

\multirow{2}{*}{Traffic} 
& MSE & 0.401 & 0.366 & \secondres{0.363}    & \boldres{0.362} & 0.364 & 0.395  \\
& MAE & 0.267 & 0.259 & \secondres{0.254}    & \boldres{0.252} & 0.255 & 0.268  \\

\midrule
\multirow{2}{*}{Weather} 
& MSE & 0.159 & 0.154 & \boldres{0.146}    & \secondres{0.148} & 0.149 & 0.152  \\
& MAE & 0.236 & 0.211 & \boldres{0.198}    & \boldres{0.198} & 0.204 & 0.215  \\
\hline
\end{tabular}
}
\label{table_mask}
\end{table}

We analyzed the sensitivity of the mask ratio parameter settings on the Electricity, Exchange, Traffic and Weather datasets.
As shown in Table \ref{table_mask}, the best performance is achieved when the mask ratio is set to 0.25.
Reasonable parameter settings can improve generalization ability and enhance the model’s robustness in unstable and missing data scenarios.

\begin{figure}[ht]
    \begin{subfigure}[b]{0.48\linewidth}
        
        \includegraphics[width=\linewidth]{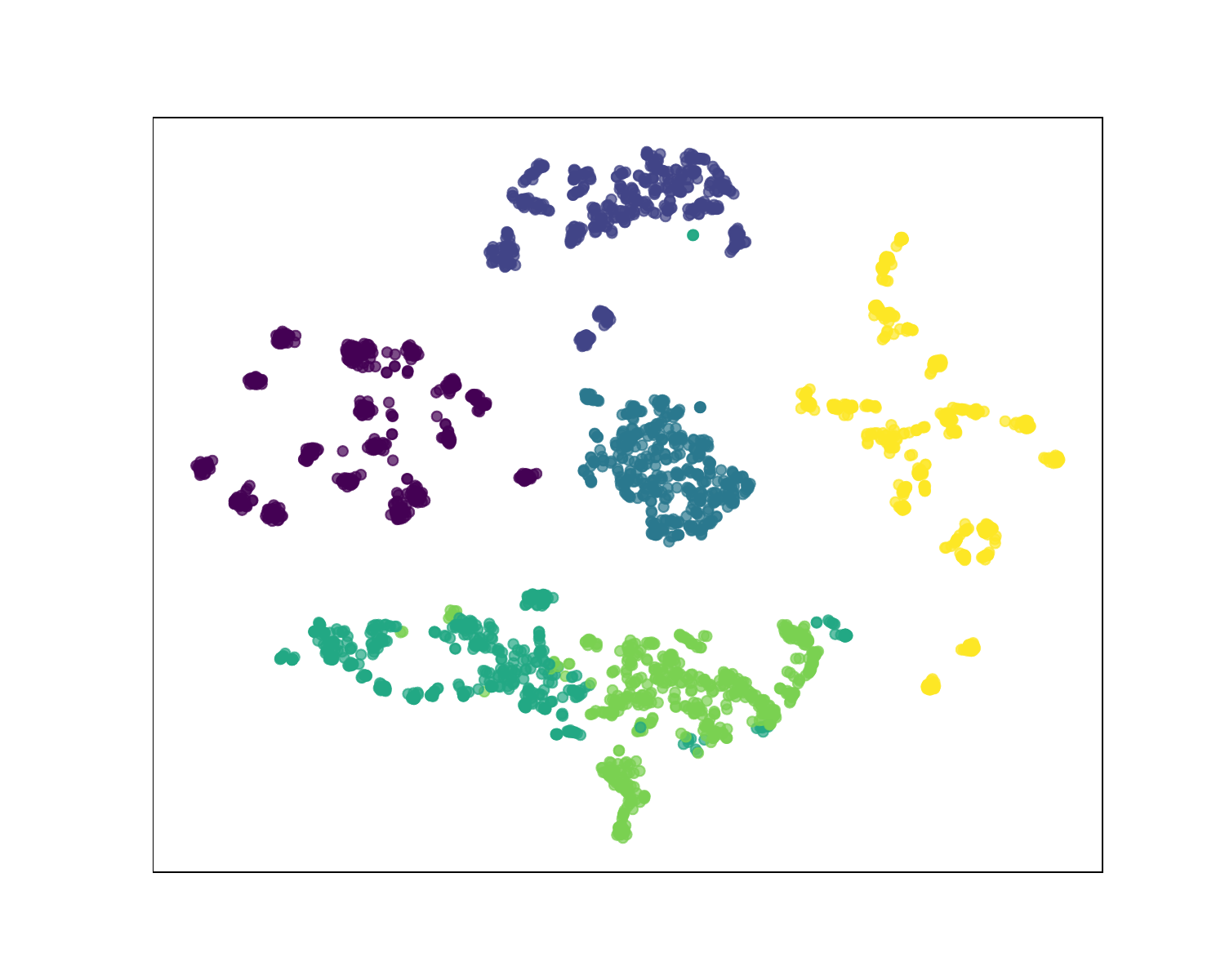}
        \caption{FusAD}
    \end{subfigure}
    \hfill
    \begin{subfigure}[b]{0.48\linewidth}
        
        \includegraphics[width=\linewidth]{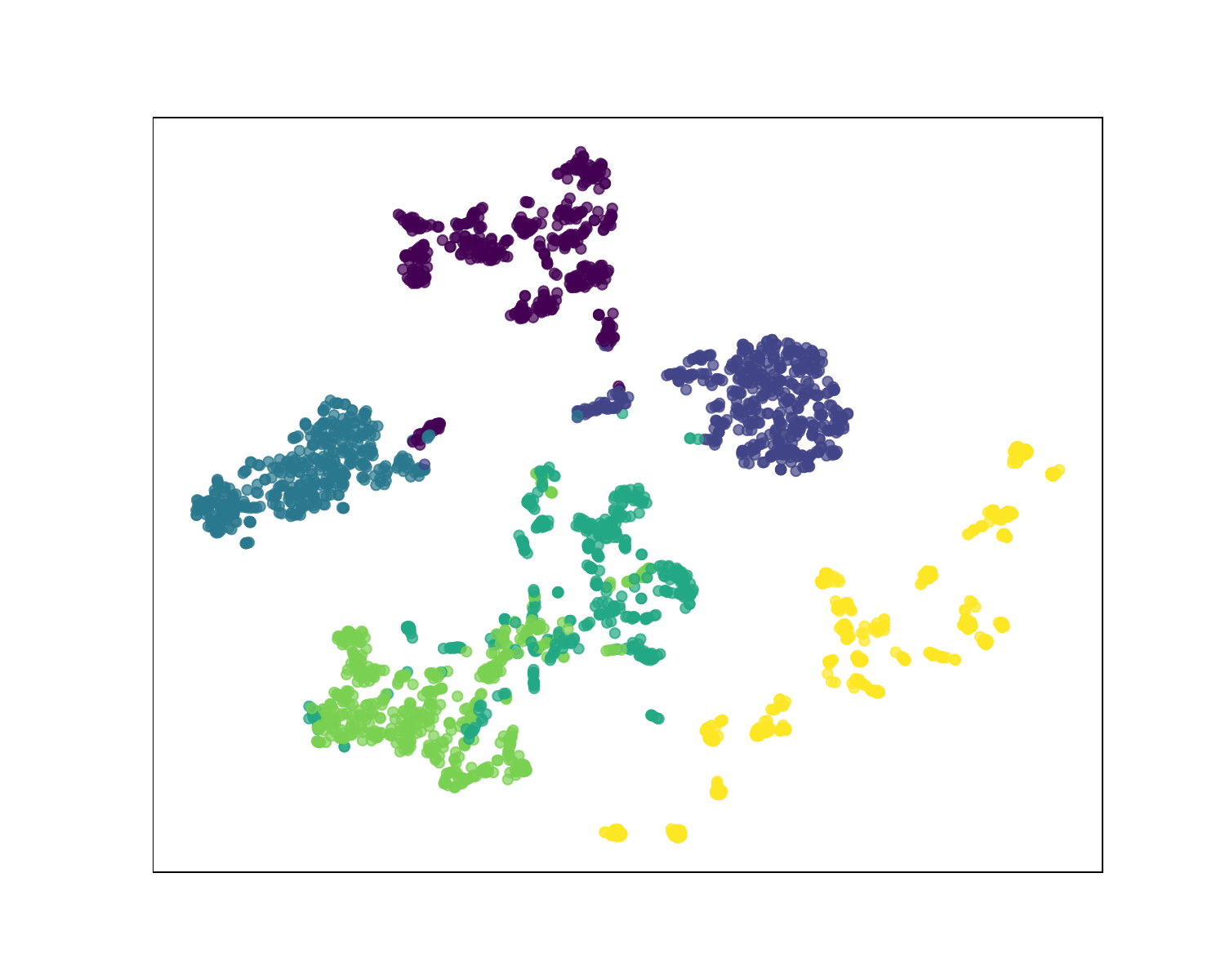}
        \caption{PatchTST}
    \end{subfigure}
    \caption{Visualization on the NATOPS classification dataset in the UEA archive.}
    \label{fig_visual_cls}
    \vspace{-0.5cm}
\end{figure}

\begin{figure}[ht]
    \begin{subfigure}[b]{0.49\linewidth}
        
        \includegraphics[width=\linewidth]{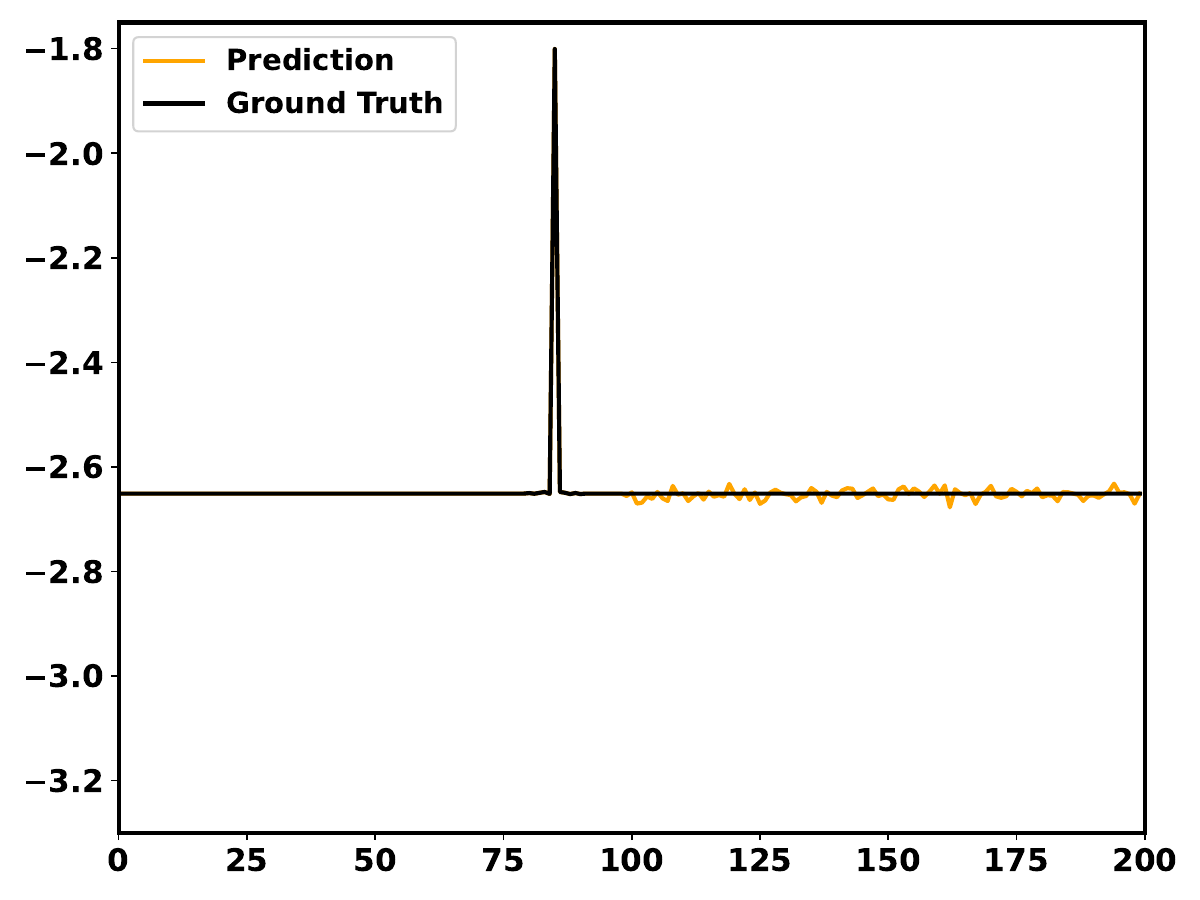}
        \caption{FusAD}
    \end{subfigure}
    \hfill
    \begin{subfigure}[b]{0.49\linewidth}
        
        \includegraphics[width=\linewidth]{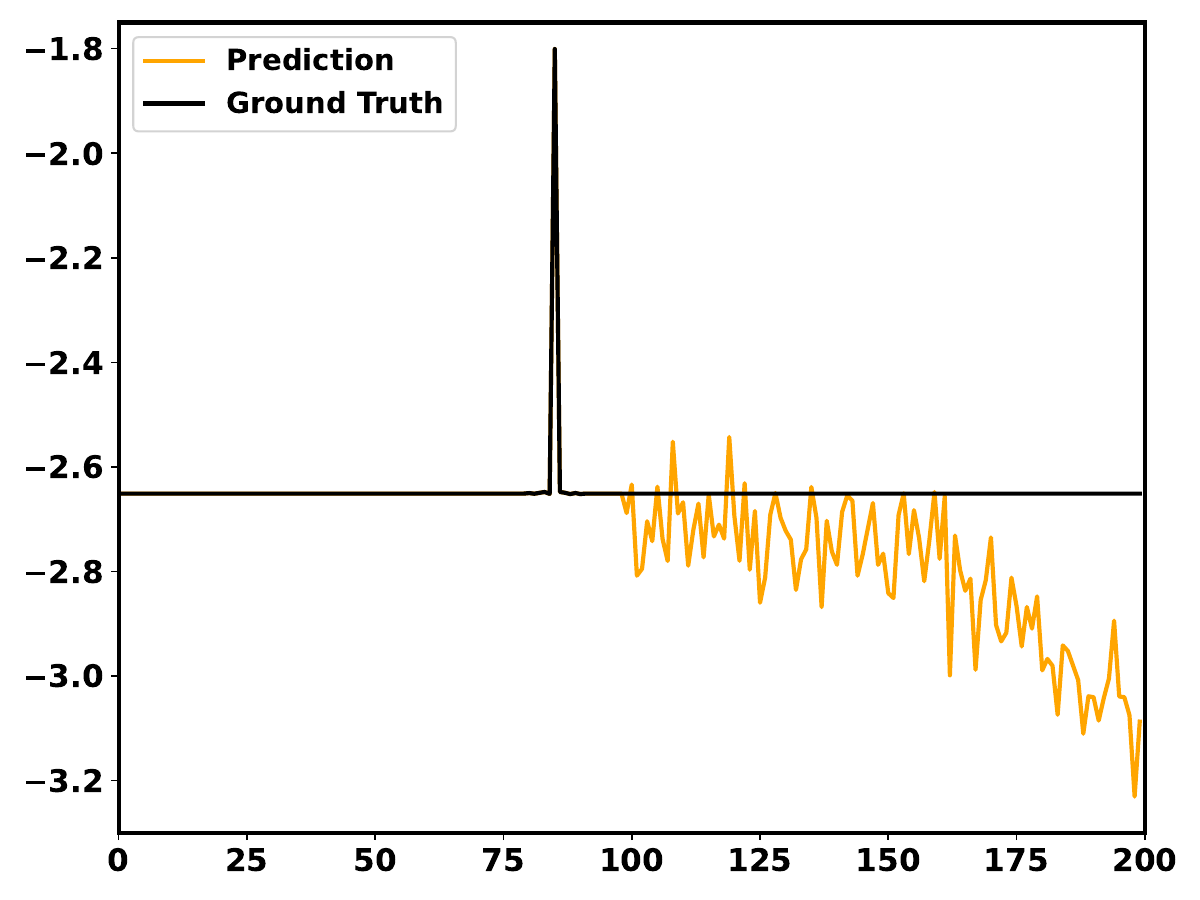}
        \caption{PatchTST}
    \end{subfigure}
    \caption{Visualization on the ETTh2 prediction dataset in EET.}
    \label{fig_visual_pre}
    \vspace{-0.4cm}
\end{figure}

\subsection{Visualization}
To further illustrate FusAD's ability to characterize learning, we visualized the classification results (Figure \ref{fig_visual_cls}) and prediction results (Figure \ref{fig_visual_pre}). The results show that FusAD places samples within the same category closer together, indicating that it better captures the intrinsic features of the data, thereby effectively representing category prototypes. Additionally, FusAD demonstrates more robust predictive capabilities compared to PatchTST, showing no drift in long-term predictions, which indicates its excellent robustness in complex data.

\section{Conclusion}
\label{conclusion}

In this work, we have presented FusAD, a novel and unified TSA framework that effectively integrates adaptive time-frequency fusion and denoising mechanisms. By jointly leveraging the strengths of Fourier and wavelet transforms, combined with an adaptive spectral filtering strategy, FusAD is capable of capturing rich multi-scale dynamic patterns while automatically suppressing noise, enabling robust feature extraction across diverse time series domains. Furthermore, the proposed information fusion module and masked pre-training enhance the model’s ability to share knowledge among tasks and improve its generalization. Extensive experiments on multiple benchmarks demonstrate that FusAD not only achieves leading performance in classification, forecasting, and anomaly detection, but also maintains high efficiency and scalability. 

\textbf{Limitation and Future Work:} Despite these advantages, some limitations remain. For instance, the framework may face scalability challenges when encountering extremely high-dimensional or highly non-stationary sequences, and adaptation to highly diverse real-world applications still requires further investigation. \textcolor{black}{In future work, we plan to explore more flexible fusion mechanisms, study model compression and acceleration for large-scale deployments, and extend FusAD to support multimodal time series, which integrate numerical series with aligned textual information \cite{liu2024time}, or irregularly sampled data, further enhancing its practicality and robustness.}

\newpage
\section*{Acknowledgments}
This work was supported in part by grants from the National Natural Science Foundation of China (62306241 \& U62576284) and from the Innovation Foundation for Doctor Dissertation of Northwestern Polytechnical University (CX2025109).

\section*{AI-Generated Content Acknowledgement}
We acknowledge the use of ChatGPT to polish the language and improve readability in the Abstract and Introduction sections of this paper. The use of this tool was limited to grammatical and stylistic refinement; the authors have verified the accuracy of the content and accept full responsibility for the final submission.

\bibliographystyle{IEEEtran}
\bibliography{IEEEabrv,mybibfile}

\begin{thebibliography}{10}
\providecommand{\url}[1]{#1}
\csname url@samestyle\endcsname
\providecommand{\newblock}{\relax}
\providecommand{\bibinfo}[2]{#2}
\providecommand{\BIBentrySTDinterwordspacing}{\spaceskip=0pt\relax}
\providecommand{\BIBentryALTinterwordstretchfactor}{4}
\providecommand{\BIBentryALTinterwordspacing}{\spaceskip=\fontdimen2\font plus
\BIBentryALTinterwordstretchfactor\fontdimen3\font minus \fontdimen4\font\relax}
\providecommand{\BIBforeignlanguage}[2]{{%
\expandafter\ifx\csname l@#1\endcsname\relax
\typeout{** WARNING: IEEEtran.bst: No hyphenation pattern has been}%
\typeout{** loaded for the language `#1'. Using the pattern for}%
\typeout{** the default language instead.}%
\else
\language=\csname l@#1\endcsname
\fi
#2}}
\providecommand{\BIBdecl}{\relax}
\BIBdecl

\bibitem{yue2022ts2vec}
Z.~Yue, Y.~Wang, J.~Duan, T.~Yang, C.~Huang, Y.~Tong, and B.~Xu, ``Ts2vec: Towards universal representation of time series,'' in \emph{Proceedings of the AAAI conference on artificial intelligence}, vol.~36, no.~8, 2022, pp. 8980--8987.

\bibitem{wu2021autoformer}
H.~Wu, J.~Xu, J.~Wang, and M.~Long, ``Autoformer: Decomposition transformers with auto-correlation for long-term series forecasting,'' \emph{Advances in neural information processing systems}, vol.~34, pp. 22\,419--22\,430, 2021.

\bibitem{chen2025impact}
H.~Chen, J.~S. Kaufman, C.~Chen, J.~Wang, A.~Maier, A.~van Dijk, N.~Slipp, J.~Rana, E.~MacIntyre, Y.~Su \emph{et~al.}, ``Impact of the 2023 wildfire smoke episodes in ontario, canada, on asthma and other health outcomes: an interrupted time-series analysis,'' \emph{CMAJ}, vol. 197, no.~17, pp. E465--E477, 2025.

\bibitem{turowski2024generating}
M.~Turowski, B.~Heidrich, L.~Weing{\"a}rtner, L.~Springer, K.~Phipps, B.~Sch{\"a}fer, R.~Mikut, and V.~Hagenmeyer, ``Generating synthetic energy time series: A review,'' \emph{Renewable and Sustainable Energy Reviews}, vol. 206, p. 114842, 2024.

\bibitem{fan2024recommender}
W.~Fan, ``Recommender systems in the era of large language models (llms),'' \emph{IEEE Transactions on Knowledge and Data Engineering}, pp. 1--20, 2024.

\bibitem{Position}
M.~Jin, Y.~Zhang, W.~Chen, K.~Zhang, Y.~Liang, B.~Yang, J.~Wang, S.~Pan, and Q.~Wen, ``Position: What can large language models tell us about time series analysis,'' in \emph{Proceedings of the 41st International Conference on Machine Learning}, 2024, pp. 22\,260--22\,276.

\bibitem{xuanomaly}
J.~Xu, H.~Wu, J.~Wang, and M.~Long, ``Anomaly transformer: Time series anomaly detection with association discrepancy,'' in \emph{International Conference on Learning Representations}, 2022.

\bibitem{liu2024time}
H.~Liu, S.~Xu, Z.~Zhao, L.~Kong, H.~Prabhakar~Kamarthi, A.~Sasanur, M.~Sharma, J.~Cui, Q.~Wen, C.~Zhang \emph{et~al.}, ``Time-mmd: Multi-domain multimodal dataset for time series analysis,'' \emph{Advances in Neural Information Processing Systems}, vol.~37, pp. 77\,888--77\,933, 2024.

\bibitem{cheng2025convtimenet}
M.~Cheng, J.~Yang, T.~Pan, Q.~Liu, Z.~Li, and S.~Wang, ``Convtimenet: A deep hierarchical fully convolutional model for multivariate time series analysis,'' in \emph{Companion Proceedings of the ACM on Web Conference 2025}, 2025, pp. 171--180.

\bibitem{lu2024trnn}
M.~Lu and X.~Xu, ``Trnn: An efficient time-series recurrent neural network for stock price prediction,'' \emph{Information Sciences}, vol. 657, p. 119951, 2024.

\bibitem{piao2025garnn}
C.~Piao, T.~Zhu, S.~E. Baldeweg, P.~Taylor, P.~Georgiou, J.~Sun, J.~Wang, and K.~Li, ``Garnn: an interpretable graph attentive recurrent neural network for predicting blood glucose levels via multivariate time series,'' \emph{Neural Networks}, p. 107229, 2025.

\bibitem{luo2024moderntcn}
D.~Luo and X.~Wang, ``Moderntcn: A modern pure convolution structure for general time series analysis,'' in \emph{The twelfth international conference on learning representations}, 2024, pp. 1--43.

\bibitem{dissanayaka2024temporal}
S.~Dissanayaka, M.~Wickramasinghe, and P.~Marasinghe, ``Temporal convolution-based hybrid model approach with representation learning for real-time acoustic anomaly detection,'' in \emph{Proceedings of the 2024 16th International Conference on Machine Learning and Computing}, 2024, pp. 218--227.

\bibitem{chen2024pioneering}
D.~Chen, S.~Liu, and L.~Yuan, ``Pioneering industrial anomaly detection with a hierarchical lstm-rola framework,'' in \emph{2024 International Joint Conference on Neural Networks (IJCNN)}.\hskip 1em plus 0.5em minus 0.4em\relax IEEE, 2024, pp. 1--8.

\bibitem{zhou2021informer}
H.~Zhou, S.~Zhang, J.~Peng, S.~Zhang, J.~Li, H.~Xiong, and W.~Zhang, ``Informer: Beyond efficient transformer for long sequence time-series forecasting,'' in \emph{Proceedings of the AAAI conference on artificial intelligence}, vol.~35, no.~12, 2021, pp. 11\,106--11\,115.

\bibitem{nietime}
Y.~Nie, N.~H. Nguyen, P.~Sinthong, and J.~Kalagnanam, ``A time series is worth 64 words: Long-term forecasting with transformers,'' in \emph{The Eleventh International Conference on Learning Representations}, 2023.

\bibitem{liuitransformer}
Y.~Liu, T.~Hu, H.~Zhang, H.~Wu, S.~Wang, L.~Ma, and M.~Long, ``itransformer: Inverted transformers are effective for time series forecasting,'' in \emph{The Twelfth International Conference on Learning Representations}, 2024.

\bibitem{wen2023transformers}
Q.~Wen, T.~Zhou, C.~Zhang, W.~Chen, Z.~Ma, J.~Yan, and L.~Sun, ``Transformers in time series: a survey,'' in \emph{Proceedings of the Thirty-Second International Joint Conference on Artificial Intelligence}, 2023, pp. 6778--6786.

\bibitem{woo2024unified}
G.~Woo, C.~Liu, A.~Kumar, C.~Xiong, S.~Savarese, and D.~Sahoo, ``Unified training of universal time series forecasting transformers,'' in \emph{International Conference on Machine Learning}.\hskip 1em plus 0.5em minus 0.4em\relax PMLR, 2024, pp. 53\,140--53\,164.

\bibitem{liu2024generative}
Z.~Liu, J.~Yang, M.~Cheng, Y.~Luo, and Z.~Li, ``Generative pretrained hierarchical transformer for time series forecasting,'' in \emph{Proceedings of the 30th ACM SIGKDD Conference on Knowledge Discovery and Data Mining}, 2024, pp. 2003--2013.

\bibitem{liang2024foundation}
Y.~Liang, H.~Wen, Y.~Nie, Y.~Jiang, M.~Jin, D.~Song, S.~Pan, and Q.~Wen, ``Foundation models for time series analysis: A tutorial and survey,'' in \emph{Proceedings of the 30th ACM SIGKDD conference on knowledge discovery and data mining}, 2024, pp. 6555--6565.

\bibitem{gao2024units}
S.~Gao, T.~Koker, O.~Queen, T.~Hartvigsen, T.~Tsiligkaridis, and M.~Zitnik, ``Units: A unified multi-task time series model,'' \emph{Advances in Neural Information Processing Systems}, vol.~37, pp. 140\,589--140\,631, 2024.

\bibitem{liu2024unitime}
X.~Liu, J.~Hu, Y.~Li, S.~Diao, Y.~Liang, B.~Hooi, and R.~Zimmermann, ``Unitime: A language-empowered unified model for cross-domain time series forecasting,'' in \emph{Proceedings of the ACM Web Conference 2024}, 2024, pp. 4095--4106.

\bibitem{wutimesnet}
H.~Wu, T.~Hu, Y.~Liu, H.~Zhou, J.~Wang, and M.~Long, ``Timesnet: Temporal 2d-variation modeling for general time series analysis,'' in \emph{The Eleventh International Conference on Learning Representations}, 2023.

\bibitem{litvnet}
C.~Li, M.~Li, and R.~Diao, ``Tvnet: A novel time series analysis method based on dynamic convolution and 3d-variation,'' in \emph{The Thirteenth International Conference on Learning Representations}, 2025.

\bibitem{zhang2024self}
K.~Zhang, Q.~Wen, C.~Zhang, R.~Cai, M.~Jin, Y.~Liu, J.~Y. Zhang, Y.~Liang, G.~Pang, D.~Song \emph{et~al.}, ``Self-supervised learning for time series analysis: Taxonomy, progress, and prospects,'' \emph{IEEE transactions on pattern analysis and machine intelligence}, 2024.

\bibitem{liu2024timesurl}
J.~Liu and S.~Chen, ``Timesurl: Self-supervised contrastive learning for universal time series representation learning,'' in \emph{Proceedings of the AAAI conference on artificial intelligence}, vol.~38, no.~12, 2024, pp. 13\,918--13\,926.

\bibitem{cheng2023timemae}
M.~Cheng, Q.~Liu, Z.~Liu, H.~Zhang, R.~Zhang, and E.~Chen, ``Timemae: Self-supervised representations of time series with decoupled masked autoencoders,'' \emph{arXiv preprint arXiv:2303.00320}, 2023.

\bibitem{liu2023koopa}
Y.~Liu, C.~Li, J.~Wang, and M.~Long, ``Koopa: Learning non-stationary time series dynamics with koopman predictors,'' \emph{Advances in neural information processing systems}, vol.~36, pp. 12\,271--12\,290, 2023.

\bibitem{dai2024ddn}
T.~Dai, B.~Wu, P.~Liu, N.~Li, X.~Yuerong, S.-T. Xia, and Z.~Zhu, ``Ddn: Dual-domain dynamic normalization for non-stationary time series forecasting,'' \emph{Advances in Neural Information Processing Systems}, vol.~37, pp. 108\,490--108\,517, 2024.

\bibitem{fan2024deep}
W.~Fan, K.~Yi, H.~Ye, Z.~Ning, Q.~Zhang, and N.~An, ``Deep frequency derivative learning for non-stationary time series forecasting,'' in \emph{Proceedings of the Thirty-Third International Joint Conference on Artificial Intelligence}, 2024, pp. 3944--3952.

\bibitem{chengrobusttsf}
H.~Cheng, Q.~Wen, Y.~Liu, and L.~Sun, ``Robusttsf: Towards theory and design of robust time series forecasting with anomalies,'' in \emph{The Twelfth International Conference on Learning Representations}, 2024.

\bibitem{ni2025timedistill}
J.~Ni, Z.~Liu, S.~Wang, M.~Jin, and W.~Jin, ``Timedistill: Efficient long-term time series forecasting with mlp via cross-architecture distillation,'' \emph{arXiv preprint arXiv:2502.15016}, 2025.

\bibitem{hansofts}
L.~Han, X.-Y. Chen, H.-J. Ye, and D.-C. Zhan, ``Softs: Efficient multivariate time series forecasting with series-core fusion,'' in \emph{The Thirty-eighth Annual Conference on Neural Information Processing Systems}, 2024.

\bibitem{zhang2023trid}
K.~Zhang, C.~Li, and Q.~Yang, ``Trid-mae: A generic pre-trained model for multivariate time series with missing values,'' in \emph{Proceedings of the 32nd ACM International Conference on Information and Knowledge Management}, 2023, pp. 3164--3173.

\bibitem{dau2019ucr}
H.~A. Dau, A.~Bagnall, K.~Kamgar, C.-C.~M. Yeh, Y.~Zhu, S.~Gharghabi, C.~A. Ratanamahatana, and E.~Keogh, ``The ucr time series archive,'' \emph{IEEE/CAA Journal of Automatica Sinica}, vol.~6, no.~6, pp. 1293--1305, 2019.

\bibitem{shumway2017arima}
R.~H. Shumway, D.~S. Stoffer, R.~H. Shumway, and D.~S. Stoffer, ``Arima models,'' \emph{Time series analysis and its applications: with R examples}, pp. 75--163, 2017.

\bibitem{tratar2016comparison}
L.~F. Tratar and E.~Strm{\v{c}}nik, ``The comparison of holt--winters method and multiple regression method: A case study,'' \emph{Energy}, vol. 109, pp. 266--276, 2016.

\bibitem{zhang2024multivariate}
D.~Zhang, J.~Gao, and X.~Li, ``Multivariate time series classification with crucial timestamps guidance,'' \emph{Expert Systems with Applications}, vol. 255, p. 124591, 2024.

\bibitem{yeh2024rpmixer}
C.-C.~M. Yeh, Y.~Fan, X.~Dai, U.~S. Saini, V.~Lai, P.~O. Aboagye, J.~Wang, H.~Chen, Y.~Zheng, Z.~Zhuang \emph{et~al.}, ``Rpmixer: Shaking up time series forecasting with random projections for large spatial-temporal data,'' in \emph{Proceedings of the 30th ACM SIGKDD Conference on Knowledge Discovery and Data Mining}, 2024, pp. 3919--3930.

\bibitem{ekambaram2023tsmixer}
V.~Ekambaram, A.~Jati, N.~Nguyen, P.~Sinthong, and J.~Kalagnanam, ``Tsmixer: Lightweight mlp-mixer model for multivariate time series forecasting,'' in \emph{Proceedings of the 29th ACM SIGKDD conference on knowledge discovery and data mining}, 2023, pp. 459--469.

\bibitem{challu2023nhits}
C.~Challu, K.~G. Olivares, B.~N. Oreshkin, F.~G. Ramirez, M.~M. Canseco, and A.~Dubrawski, ``Nhits: Neural hierarchical interpolation for time series forecasting,'' in \emph{Proceedings of the AAAI conference on artificial intelligence}, vol.~37, no.~6, 2023, pp. 6989--6997.

\bibitem{zeng2023transformers}
A.~Zeng, M.~Chen, L.~Zhang, and Q.~Xu, ``Are transformers effective for time series forecasting?'' in \emph{Proceedings of the AAAI conference on artificial intelligence}, vol.~37, no.~9, 2023, pp. 11\,121--11\,128.

\bibitem{chen2024pathformer}
P.~Chen, Y.~Zhang, Y.~Cheng, Y.~Shu, Y.~Wang, Q.~Wen, B.~Yang, and C.~Guo, ``Pathformer: Multi-scale transformers with adaptive pathways for time series forecasting,'' in \emph{International Conference on Learning Representations}, 2024.

\bibitem{liupyraformer}
S.~Liu, H.~Yu, C.~Liao, J.~Li, W.~Lin, A.~X. Liu, and S.~Dustdar, ``Pyraformer: Low-complexity pyramidal attention for long-range time series modeling and forecasting,'' in \emph{International Conference on Learning Representations}, 2022.

\bibitem{fraikin2024t}
A.~Fraikin, A.~Bennetot, and S.~Allassonni{\`e}re, ``T-rep: Representation learning for time series using time-embeddings,'' in \emph{The Twelfth International Conference on Learning Representations}, 2024.

\bibitem{zhou2023one}
T.~Zhou, P.~Niu, L.~Sun, R.~Jin \emph{et~al.}, ``One fits all: Power general time series analysis by pretrained lm,'' \emph{Advances in neural information processing systems}, vol.~36, pp. 43\,322--43\,355, 2023.

\bibitem{zamanzadeh2024deep}
Z.~Zamanzadeh~Darban, G.~I. Webb, S.~Pan, C.~Aggarwal, and M.~Salehi, ``Deep learning for time series anomaly detection: A survey,'' \emph{ACM Computing Surveys}, vol.~57, no.~1, pp. 1--42, 2024.

\bibitem{zhang2025beyond}
Q.~Zhang, P.~Yang, H.~Wen, X.~Li, H.~Wang, F.~Sun, Z.~Song, Z.~Lai, R.~Ma, R.~Han \emph{et~al.}, ``Beyond the time domain: Recent advances on frequency transforms in time series analysis,'' \emph{arXiv preprint arXiv:2504.07099}, 2025.

\bibitem{dong2024timesiam}
J.~Dong, H.~Wu, Y.~Wang, Y.~Qiu, L.~Zhang, J.~Wang, and M.~Long, ``Timesiam: a pre-training framework for siamese time-series modeling,'' in \emph{Proceedings of the 41st International Conference on Machine Learning}, 2024, pp. 11\,412--11\,436.

\bibitem{zhou2022film}
T.~Zhou, Z.~Ma, Q.~Wen, L.~Sun, T.~Yao, W.~Yin, R.~Jin \emph{et~al.}, ``Film: Frequency improved legendre memory model for long-term time series forecasting,'' \emph{Advances in neural information processing systems}, vol.~35, pp. 12\,677--12\,690, 2022.

\bibitem{liu2022non}
Y.~Liu, H.~Wu, J.~Wang, and M.~Long, ``Non-stationary transformers: Exploring the stationarity in time series forecasting,'' \emph{Advances in neural information processing systems}, vol.~35, pp. 9881--9893, 2022.

\bibitem{yefrequency}
W.~Ye, S.~Deng, Q.~Zou, and N.~Gui, ``Frequency adaptive normalization for non-stationary time series forecasting,'' in \emph{The Thirty-eighth Annual Conference on Neural Information Processing Systems}, 2024.

\bibitem{zhang2025waveletmixer}
Z.~Zhang, T.~D. Pham, Y.~An, N.~P. Doan, M.~Alsharari, V.-H. Tran, A.-T. Hoang, H.~Vandierendonck, and S.~T. Mai, ``Waveletmixer: a multi-resolution wavelets based mlp-mixer for multivariate long-term time series forecasting,'' in \emph{Proceedings of the AAAI Conference on Artificial Intelligence}, vol.~39, no.~21, 2025, pp. 22\,741--22\,749.

\bibitem{murad2025wpmixer}
M.~M.~N. Murad, M.~Aktukmak, and Y.~Yilmaz, ``Wpmixer: Efficient multi-resolution mixing for long-term time series forecasting,'' in \emph{Proceedings of the AAAI Conference on Artificial Intelligence}, vol.~39, no.~18, 2025, pp. 19\,581--19\,588.

\bibitem{xufits}
Z.~Xu, A.~Zeng, and Q.~Xu, ``Fits: Modeling time series with $10 k $ parameters,'' in \emph{The Twelfth International Conference on Learning Representations}, 2024.

\bibitem{wu2019graph}
Z.~Wu, S.~Pan, G.~Long, J.~Jiang, and C.~Zhang, ``Graph wavenet for deep spatial-temporal graph modeling,'' in \emph{Proceedings of the 28th International Joint Conference on Artificial Intelligence}, 2019, pp. 1907--1913.

\bibitem{zhou2024denoising}
S.~Zhou, D.~Zha, X.~Shen, X.~Huang, R.~Zhang, and K.~Chung, ``Denoising-aware contrastive learning for noisy time series,'' in \emph{33rd International Joint Conference on Artificial Intelligence, IJCAI 2024}.\hskip 1em plus 0.5em minus 0.4em\relax International Joint Conferences on Artificial Intelligence, 2024, pp. 5644--5652.

\bibitem{liu2024wftnet}
P.~Liu, B.~Wu, N.~Li, T.~Dai, F.~Lei, J.~Bao, Y.~Jiang, and S.-T. Xia, ``Wftnet: Exploiting global and local periodicity in long-term time series forecasting,'' in \emph{ICASSP 2024-2024 IEEE International Conference on Acoustics, Speech and Signal Processing (ICASSP)}.\hskip 1em plus 0.5em minus 0.4em\relax IEEE, 2024, pp. 5960--5964.

\bibitem{bagnall2018uea}
A.~Bagnall, H.~A. Dau, J.~Lines, M.~Flynn, J.~Large, A.~Bostrom, P.~Southam, and E.~Keogh, ``The uea multivariate time series classification archive, 2018,'' \emph{arXiv preprint arXiv:1811.00075}, 2018.

\bibitem{eldele2021time}
E.~Eldele, M.~Ragab, Z.~Chen, M.~Wu, C.~K. Kwoh, X.~Li, and C.~Guan, ``Time-series representation learning via temporal and contextual contrasting,'' in \emph{Proceedings of the Thirtieth International Joint Conference on Artificial Intelligence}, 2021.

\bibitem{pieper2023self}
F.~Pieper, K.~Ditschuneit, M.~Genzel, A.~Lindt, and J.~Otterbach, ``Self-distilled representation learning for time series,'' \emph{arXiv preprint arXiv:2311.11335}, 2023.

\bibitem{tonekaboniunsupervised}
S.~Tonekaboni, D.~Eytan, and A.~Goldenberg, ``Unsupervised representation learning for time series with temporal neighborhood coding,'' in \emph{International Conference on Learning Representations}, 2021.

\bibitem{wang2024graph}
Y.~Wang, Y.~Xu, J.~Yang, M.~Wu, X.~Li, L.~Xie, and Z.~Chen, ``Graph-aware contrasting for multivariate time-series classification,'' in \emph{Proceedings of the AAAI conference on artificial intelligence}, vol.~38, no.~14, 2024, pp. 15\,725--15\,734.

\bibitem{tian2025frera}
T.~Tian, C.~Miao, and H.~Qian, ``Frera: A frequency-refined augmentation for contrastive learning on time series classification,'' \emph{arXiv preprint arXiv:2505.23181}, 2025.

\bibitem{hossin2015review}
M.~Hossin and M.~N. Sulaiman, ``A review on evaluation metrics for data classification evaluations,'' \emph{International journal of data mining \& knowledge management process}, vol.~5, no.~2, p.~1, 2015.

\bibitem{demvsar2006statistical}
J.~Dem{\v{s}}ar, ``Statistical comparisons of classifiers over multiple data sets,'' \emph{Journal of Machine learning research}, vol.~7, no. Jan, pp. 1--30, 2006.

\bibitem{lai2018modeling}
G.~Lai, W.-C. Chang, Y.~Yang, and H.~Liu, ``Modeling long-and short-term temporal patterns with deep neural networks,'' in \emph{The 41st international ACM SIGIR conference on research \& development in information retrieval}, 2018, pp. 95--104.

\bibitem{zhang2023crossformer}
Y.~Zhang and J.~Yan, ``Crossformer: Transformer utilizing cross-dimension dependency for multivariate time series forecasting,'' in \emph{The eleventh international conference on learning representations}, 2023.

\bibitem{zhou2022fedformer}
T.~Zhou, Z.~Ma, Q.~Wen, X.~Wang, L.~Sun, and R.~Jin, ``Fedformer: Frequency enhanced decomposed transformer for long-term series forecasting,'' in \emph{International conference on machine learning}.\hskip 1em plus 0.5em minus 0.4em\relax PMLR, 2022, pp. 27\,268--27\,286.

\bibitem{li2023revisiting}
Z.~Li, S.~Qi, Y.~Li, and Z.~Xu, ``Revisiting long-term time series forecasting: An investigation on linear mapping,'' \emph{arXiv preprint arXiv:2305.10721}, 2023.

\bibitem{wu2025affirm}
Y.~Wu, X.~Meng, H.~Hu, J.~Zhang, Y.~Dong, and D.~Lu, ``Affirm: Interactive mamba with adaptive fourier filters for long-term time series forecasting,'' in \emph{Proceedings of the AAAI Conference on Artificial Intelligence}, vol.~39, no.~20, 2025, pp. 21\,599--21\,607.

\bibitem{hu2025adaptive}
Y.~Hu, P.~Liu, P.~Zhu, D.~Cheng, and T.~Dai, ``Adaptive multi-scale decomposition framework for time series forecasting,'' in \emph{Proceedings of the AAAI Conference on Artificial Intelligence}, vol.~39, no.~16, 2025, pp. 17\,359--17\,367.

\bibitem{su2019robust}
Y.~Su, Y.~Zhao, C.~Niu, R.~Liu, W.~Sun, and D.~Pei, ``Robust anomaly detection for multivariate time series through stochastic recurrent neural network,'' in \emph{Proceedings of the 25th ACM SIGKDD international conference on knowledge discovery \& data mining}, 2019, pp. 2828--2837.

\bibitem{hundman2018detecting}
K.~Hundman, V.~Constantinou, C.~Laporte, I.~Colwell, and T.~Soderstrom, ``Detecting spacecraft anomalies using lstms and nonparametric dynamic thresholding,'' in \emph{Proceedings of the 24th ACM SIGKDD international conference on knowledge discovery \& data mining}, 2018, pp. 387--395.

\bibitem{mathur2016swat}
A.~P. Mathur and N.~O. Tippenhauer, ``Swat: A water treatment testbed for research and training on ics security,'' in \emph{2016 international workshop on cyber-physical systems for smart water networks (CySWater)}.\hskip 1em plus 0.5em minus 0.4em\relax IEEE, 2016, pp. 31--36.

\bibitem{abdulaal2021practical}
A.~Abdulaal, Z.~Liu, and T.~Lancewicki, ``Practical approach to asynchronous multivariate time series anomaly detection and localization,'' in \emph{Proceedings of the 27th ACM SIGKDD conference on knowledge discovery \& data mining}, 2021, pp. 2485--2494.

\bibitem{li2024tsinr}
M.~Li, K.~Liu, H.~Chen, J.~Bu, H.~Wang, and H.~Wang, ``Tsinr: Capturing temporal continuity via implicit neural representations for time series anomaly detection,'' \emph{arXiv preprint arXiv:2411.11641}, 2024.

\end{thebibliography}

\end{document}